\documentclass{article}
\usepackage{amsmath}
\usepackage{amssymb}
\usepackage{mathtools}
\usepackage[a4paper]{geometry}
\usepackage{rotating}
\usepackage{xcolor,colortbl}
\usepackage{natbib}
\usepackage{hyperref}
\usepackage{booktabs}
\usepackage[font=small]{caption}
\usepackage{algorithm,algpseudocode}
\usepackage{changepage}

\usepackage{graphicx}

\newcolumntype{P}[1]{>{\centering\arraybackslash}p{#1}}
\newcolumntype{M}[1]{>{\centering\arraybackslash}m{#1}}

\DeclareMathOperator*{\argmax}{arg\,max}

\begin{document}

\title{A Unifying Framework for Reinforcement Learning and Planning}
\author{Thomas M. Moerland$^{1}$, Joost Broekens$^1$, Aske Plaat$^1$ and Catholijn M. Jonker$^{1,2}$ \\
$^1$ Leiden Institute of Advanced Computer Science, Leiden University, The Netherlands \\
$^2$ Interactive Intelligence, TU Delft, The Netherlands
}
\date{}

\maketitle

\tableofcontents

\abstract{Sequential decision making, commonly formalized as optimization of a Markov Decision Process, is a key challenge in artificial intelligence. Two successful approaches to MDP optimization are {\it reinforcement learning} and {\it planning}, which both largely have their own research communities. However, if both research fields solve the same problem, then we might be able to disentangle the common factors in their solution approaches. Therefore, this paper presents a unifying algorithmic framework for reinforcement learning and planning (FRAP), which identifies underlying dimensions on which MDP planning and learning algorithms have to decide. At the end of the paper, we compare a variety of well-known planning, model-free and model-based RL algorithms along these dimensions. Altogether, the framework may help provide deeper insight in the algorithmic design space of planning and reinforcement learning.}

\vspace{0.5cm}

\noindent {\bf Keywords}: Planning, reinforcement learning, model-based reinforcement learning, framework, overview, synthesis. 

\section{Introduction} \label{sec_introduction}
Sequential decision making is a key challenge in artificial intelligence (AI) research. The problem, commonly formalized as a Markov Decision Process (MDP) \citep{bellman1954theory,puterman2014markov}, has been studied in different research fields. The two prime research directions are {\it reinforcement learning} (RL) \citep{sutton2018reinforcement}, a subfield of machine learning, and {\it planning} (also known as {\it search}), of which the discrete and continuous variants have been studied in the fields of artificial intelligence \citep{russell2016artificial} and control theory \citep{bertsekas1995dynamic}, respectively. Departing from different assumptions both fields have largely developed their own methodology, which has cross-pollinated in the field of {\it model-based reinforcement learning} \citep{sutton1990integrated,moerland2020model,hamrick2019analogues,plaat2021high}. 

However, a unified view on both fields, including how their approaches overlap or differ, lacks in literature. For example, the classic AI textbook by \citet{russell2016artificial} discusses (heuristic) search methods in Chapters 3, 4, 10 and 11, while reinforcement learning methodology is separately discussed in Chapter 21. Similarly, the classic RL textbook by \citet{sutton2018reinforcement} does discuss a variety of the topics in our framework, but never summarizes these as a single algorithmic space. Moreover, while the book does extensively discuss the relation between reinforcement learning and dynamic programming methods, it does not focus on the relation with the many other branches of planning literature. Therefore, this paper introduces a Framework for Reinforcement learning and Planning (FRAP) (Table \ref{table_framework}), which attempts to identify the underlying algorithmic space shared by RL and MDP planning algorithms. We show that a wide range of algorithms, from Q-learning \citep{watkins1992q} to Dynamic Programming \citep{bellman1954theory} to A$^\star$ \citep{hart1968formal}, fit the framework, simply making different decisions on a number of subdimensions of the framework (Table \ref{table_overview}). 

We need to warn experienced readers that many of the individual topics in the paper will be familiar to them. However, the main contribution of this paper is not the discussion of these ideas themselves, but in the {\it systematic structuring} of these ideas into a single algorithmic space (Algorithm \ref{alg_frap}). Experienced readers may therefore skim over some sections more quickly, and only focus on the bigger integrative message. As a second contribution, we hope the paper points researchers from one of both fields towards relevant literature from the other field, thereby stimulating cross-pollination. Third, we note that the framework is equally useful for researchers from model-free RL, since to the best of our knowledge `a framework for reinforcement learning' does not exist in literature either (`a framework for planning' does, see Related Work). Finally, we hope the paper may also serves an educational purpose, for example for students in a university course, by putting algorithms that are often presented in different courses into a single perspective. 

We also need to clearly demarcate what literature we do and do not include. First of all, planning and reinforcement learning are huge research fields, and the present paper is definitely {\it not} a systematic survey of both fields (which would likely require multiple books, not a single article). Instead, we focus on the core ideas in the joint algorithmic space and discuss characteristic, well-known algorithms to illustrate these key ideas. For the planning side of the literature, we exclusively focus on planning algorithms that search for {\it optimal behaviour} in an MDP formulation, which for example excludes all non-MDP planning methods, as well as `planning as satistifiability' approaches, which attempt to verify whether a path from start to goal exists at all \citep{kautz1992planning,kautz2006satplan}. For the reinforcement learning side of the literature, we do not focus on approaches that treat the MDP formulation as a {\it black-box optimization problem}, such as evolutionary algorithms \citep{moriarty1999evolutionary}, simulated annealing \citep{atiya2003reinforcement} or the cross-entropy method \citep{mannor2003cross}. While these approaches can be successful \citep{salimans2017evolution}, they typically only require access to an evaluation function, and do not use MDP specific characteristics in their solution (on which our framework is built). 

The remainder of this article is organized as follows. After discussing Related Work (Sec. \ref{sec_relatedwork}), we first formally introduce the MDP optimization setting (Sec. \ref{sec_mdp}), the way we may get access to the MDP (Sec. \ref{sec_model_types}), and give definitions of planning and reinforcement learning (Sec. \ref{sec_plan_rl_definitions}). The next section provides brief overviews of literature in planning (Sec. \ref{sec_planning}) and reinforcement learning (Sec. \ref{sec_rl}). Together, Sections \ref{sec_problem_def} and \ref{sec_literature} should establish common ground to build the framework upon. The main contribution of this paper, the framework, is presented in Section \ref{sec_unifying_view}, where we  systematically discuss each consideration in the algorithmic space. Finally, Section \ref{sec_comparison} illustrates the applicability of the framework, by comparing a range of planning and reinforcement learning algorithms along the framework dimensions, and identifying interesting directions for future work.

\section{Related Work} \label{sec_relatedwork}
The basis for a framework approach to planning (and reinforcement learning) is the FIND-and-REVISE scheme by \citet{bonet2003faster}. FIND-and-REVISE specifies a general procedure for asynchronous value iteration, where we first {\it find} a new node that requires updating, and subsequently {\it revise} the value estimate of that node based on interaction with the MDP. Our framework follows as similar pattern, where we repeatedly find a new state (a root that requires updating), find interesting subsequent states to compute an improved value estimate for this state, and subsequently use this estimate to improve the solution. Our framework is also partially inspired by the reinforcement learning textbook of \citet{sutton2018reinforcement}, which provides an unified view on the back-up patterns in planning and reinforcement learning (regarding their depth and width), and thereby an integrated view on dynamic programming and reinforcement learning methodology. Similar ideas return in our framework, but we extend them with several additional dimensions, and to a wide variety of other planning literature. 

However, the main inspiration of our work is {\it trial-based heuristic tree search} (THTS) \citep{keller2015anytime,keller2013trial}, a framework that subsumed several planning algorithms, like Dynamic Programming \citep{bellman1954theory}, MCTS \citep{kocsis2006bandit} and heuristic search \citep{pearl1984heuristics} methods. THTS shows that a variety of planning algorithms can indeed be unified in the same algorithmic space, which we believe provided a lot of insight in the commonalities of these algorithms. Our present framework can be seen as an extension and modification of these ideas to also incorporate literature from the reinforcement learning community. Compared to THTS, we first of all add several new categories to the framework, such as `solution representation' and `update of the solution', to accommodate for the various ways in which planning and RL methods differ in the way they store and update the outcome of their back-ups. Second, THTS purely focused on the online planning setting, while we incorporate a new dimension `set root state' that also allows for different prioritization schemes in offline planning and learning. Third, we make several smaller adjustments and extensions, such as splitting up the back-up dimension in several subdimensions, and using a different definition of the concept of a trial (which we define as a single forward sequence of states and actions), which allows us to bound the computational effort per trial. This also leads to a new `budget per root' dimension in the framework, which now specifies the number of trials (width) of the unfolded subtree in the local solution. We nevertheless invite the reader to also read the THTS papers, since they are a useful companion to the present paper.

\section{Definitions} \label{sec_problem_def}
In sequential decision-making, formalized as Markov Decision Process optimization, we are interested in the following problem: given a (sequence of) state(s), what next action is best to choose, based on the criterion of highest cumulative pay-off in the future. More formally, we aim for {\it context-dependent action prioritization based on a (discounted) cumulative reward criterion}. This is a core challenge in artificial intelligence research, as it contains the key elements of the world: there is sensory information about the environment (states), we can influence that environment through actions, and there is some notion of what is preferable, now and in the future. The formulation can deal with a wide variety of well-known problem instances, like path planning, robotic manipulation, game playing and autonomous driving.

\subsection{Markov Decision Process} \label{sec_mdp}
The formal definition of a {\it Markov Decision Process} (MDP) \citep{puterman2014markov} is a tuple $\mathcal{M} = \{\mathcal{S},\mathcal{A},\mathcal{T},\mathcal{R},\gamma, p_0(s)\}$.  The environment consists of a {\it transition function} $\mathcal{T}:\mathcal{S} \times \mathcal{A} \to p(\mathcal{S})$ and a {\it reward function} $\mathcal{R}: \mathcal{S} \times \mathcal{A} \times \mathcal{S} \to \mathbb{R}$. At each timestep $t$ we observe some state $s_t \in \mathcal{S}$ and pick an action $a_t \in \mathcal{A}$. Then, the environment returns a next state $s_{t+1} \sim \mathcal{T}(s_{t+1}|s_t,a_t)$ and associated scalar reward $r_t = \mathcal{R}(s_t,a_t,s_{t+1})$. The first state is sampled from the initial state distribution $p_0(s)$, while $\gamma \in [0,1]$ denotes a discount parameter.

The state space can either have an atomic, factorized, or structured form \citep{russell2016artificial}. {\it Atomic} state spaces treat each state as a separate, discrete entity, without the specification of any additional relation between states. In contrast, factorized states consist of a vector of attributes, which thereby provide a relation between different states (i.e., the attributes of states may partially overlap). Factorized state spaces allow for {\it generalization} between states, an important feature of learning algorithms. Finally, {\it structured} state spaces consist of factorized states with additional structure beyond simple discrete or continuous values, for example in the form of a symbolic language. In this work, we primarily focus on settings with atomic or factorized states.

The agent acts in the environment according to a {\it policy} $\pi: \mathcal{S} \to p(\mathcal{A})$. In the search community, a policy is also known as a {\it contingency plan} or {\it strategy} \citep{russell2016artificial}. By repeatedly selecting actions and transitioning to a next state, we can sample a {\it trace} through the environment. The {\it cumulative return} of the trace is denoted by: $J_t = \sum_{k=0}^K (\gamma)^k \cdot r_{t+k}$, for a trace of length $K$. For $K=\infty$ we call this the infinite-horizon return. The action-value function $Q^\pi(s,a)$ is defined as the expectation of this cumulative return given a particular policy $\pi$:

\begin{equation} 
Q^\pi(s,a) \dot{=} \mathbb{E}_{\pi,\mathcal{T}} \Bigg[ \sum_{k=0}^K (\gamma)^k \cdot r_{t+k}  \Big| s_t=s, a_t=a \Bigg] \label{eq_cum_reward}
\end{equation}

This equation can be written in a recursive form, better known as the {\it Bellman equation}:
 
\begin{align}
Q^\pi(s,a) &= \mathbb{E}_{s' \sim \mathcal{T}(\cdot|s,a)} \Bigg[ \mathcal{R}(s,a,s') + \gamma \hspace{0.1cm}\mathbb{E}_{a' \sim \pi(\cdot|s')}\Big[Q^\pi(s',a') \Big]  \Bigg] \label{eq_bellman}
\end{align}

Our goal is to find a policy $\pi$ that maximizes our expected return $Q^\pi(s,a)$:

\begin{equation} 
\pi^\star = \argmax_\pi Q^\pi(s,a) \label{eq_mdp_opt}
\end{equation}

In the planning and control literature, the above problem is typically formulated as a cost {\it minimization} problem \citep{bellman1957markovian}. That formulation is interchangeable with our presentation by negating the reward function. The formulation also contains {\it stochastic shortest path} (SSP) problems \citep{bertsekas1991analysis}, which are a common setting in the planning literature. SSP problems are MDP specifications with negative rewards on all transitions and particular terminal goal states, where we attempt to reach the goal with as little cost as possible. The MDP specification induces a graph, which is in the planning community commonly referred to as an {\it AND-OR graph}: we repeatedly need to choose between actions (OR), and then take the expectation over the next states (AND). In a search tree these two operations are sometimes referred to as {\it decision nodes} (OR) and {\it chance nodes} (AND), respectively.

\begin{figure}[t]
  \centering
      \includegraphics[width = 0.60\textwidth]{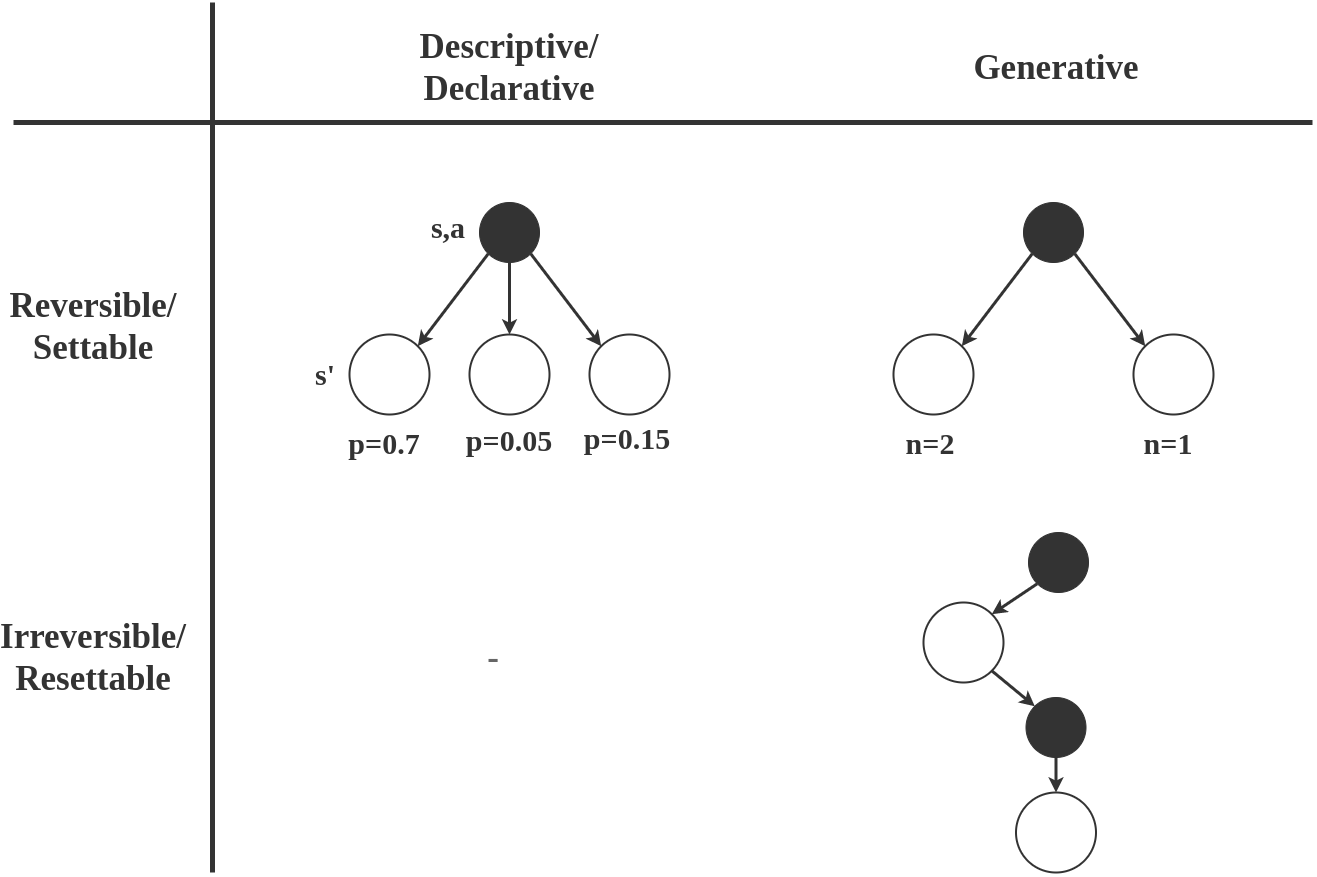}
  \caption{Illustration of different types of access to the MDP transition dynamics. {\bf Rows}: We may either have {\it reversible/settable} access to the MDP dynamics, in which case we can query the MDP on any desired state, or {\it irreversible/resettable} access to the MDP, in which case we have to make the next query at the resulting state, or we can reset to a state from the initial state distribution. Any type of reversible/settable access to the MDP is usually called a (known) {\it model}. {\bf Columns}: On each query to the MDP dynamics, we may either get access to the full distribution of possible next states ({\it descriptive}/{\it declarative} access), or only get a single sample from this distribution ({\it generative} access). Note that we could theoretically think of irreversible descriptive access, in which we do see the probabilities but need to continue from the next state, but we are unaware of such a model in practice.}   
   \label{fig_model_types}
\end{figure}

\subsection{Access to the MDP dynamics} \label{sec_model_types}
A crucial aspect in MDP optimization is the way we can interact with the MDP, i.e., the {\it type of access} we have to the transition and reward function. We will here focus on the type of access to the transition function, since the type of access to the reward usually mimics the type of access to the transition function. All MDP algorithms at some point {\it query} the MDP transition function at a particular state-action pair $(s,a)$, and get information back about the possible next state(s) $s'$ and associated reward $\mathcal{R}(s,a,s')$. However, there are differences in the {\it order} in which we can make queries, and in the {\it type of information} we get back after a query \citep{kearns2002sparse,keller2013trial}.   

Regarding the first consideration, reinforcement learning methods often assume we need to make our next query at the state that resulted from our last query, i.e., we have to move forward (similar to the way humans interact with the real world). We propose to call this {\it irreversible} access to the MDP, since we cannot revert a particular action. In practice, RL approaches often assume that we can reset at any particular moment to a state sampled from the initial state distribution, so we may also call this {\it resettable} access to the MDP. In contrast, planning methods often assume we can query the MDP dynamics in any preferred order of state-action pairs, i.e., we can {\it set} the query to any state we like. This property also allows us to repeatedly plan forward from the same state (like humans plan in their mind), which we therefore propose to call {\it reversible} access to the MDP dynamics. The distinction between reversible/settable and irreversible/resettable access is visualized in the rows of Figure \ref{fig_model_types}. Reversible/settable access to the MDP dynamics is usually referred to as a (known) {\it model}. 

\begin{quote} \centering \it 
A model is a type of access to the MDP dynamics that can be queried in any preferred order of state-action pairs.
\end{quote}

\noindent A second important distinction concerns the type of information we get about the possible next states. A {\it descriptive/declarative} model provides us the full probabilities of each possible next state, i.e., the entire distribution of $\mathcal{T}(s'|s,a)$, which allows us to fully evaluate the expectation over the dynamics in the Bellman equation (Eq. \ref{eq_bellman}). In contrast, {\it generative} access only provides us with a sample from the next state distribution, without access to the true underlying probabilities (we may of course approximate the expectation in Eq. \ref{eq_bellman} through repeated sampling). These two options are displayed in the columns of Fig. \ref{fig_model_types}). 

Together, the two considerations lead to three types of access to the MDP dynamics, as shown in the cells of Figure \ref{fig_model_types}. Reversible descriptive access (top-left) is for example used in Value Iteration \citep{bellman1957markovian}, reversible generative access (top-right) is used in Monte Carlo Tree Search \citep{kocsis2006bandit}, while irreversible generative access (bottom-right) is used in Q-learning \citep{watkins1992q}. The combination of irreversible and descriptive access, in the bottom-left of Figure \ref{fig_model_types}), is theoretically possible, but to our knowledge does not occur in practice. Note that there is also a natural ordering in these types of MDP access: reversible descriptive access gives most information and freedom, followed by reversible generative access (since we can always sample from distributional access), and then followed by irreversible generative access (since we can always restrict the order of sampling ourselves). However, the difficulty to obtain a particular type of access follows the opposite pattern: descriptive models are typically hardest to obtain, while a irreversible generative access is by definition available through real-world interaction.  

\subsection{Definitions of planning and reinforcement learning} \label{sec_plan_rl_definitions}
We are now ready to give formal definitions of MDP planning and reinforcement learning. While there are various definitions of both fields in literature \citep{sutton2018reinforcement,russell2016artificial}, these are typically not specific enough to discriminate planning from reinforcement learning. One possible distinction is based on the {\it type of access} to the MDP dynamics: planning approaches had settable/reversible access to the dynamics (`a known model'), while reinforcement learning approaches had irreversible access (`an unknown model'). However, there is a second possible distinction, based on the {\it coverage or storage of the solution}. This distinction seems known to many researchers, but is seldomly expicitly discussed in research papers. On the one hand, planning methods tend to use {\it local} solution representations: the solution is only stored temporarily, and usually valid for only a subset of all states (for example repeatedly simulating forward from a current state). In contrast, reinforcement learning approaches tend to use a {\it global} solution: a permanent storage of the solution which is typically valid for all possible states.

\begin{quote} \centering \it 
A local solution temporarily stores solution estimates for a subset of all states. 
\end{quote}

\begin{quote} \centering \it 
A global solution permanently stores solution estimates for all states.
\end{quote}

\noindent The focus of RL methods on global solutions is easy to understand: without a model we cannot repeatedly simulate forward from the same state, and therefore our best bet is to store a solution for all possible states (we can never build a local solution beyond size one, since we have to move forward). The global solutions that we gradually update are typically referred to as {\it learned} solutions, which connects reinforcement learning to the broader machine learning literature.  

Interestingly, our two possible distinctions between planning and reinforcement learning (model versus no model, and local versus global solution) do not always agree. For example, both Value Iteration \citep{bellman1966dynamic} and AlphaZero \citep{silver2018general} combine a global solution (which would make it reinforcement learning) with a model (which would make it planning). Indeed, Dynamic Programming has long been considered a bridging technique between planning and reinforcement learning. We propose to solve this issue by considering these borderline cases as {\it model-based reinforcement learning} \citep{samuel1967some,sutton1990integrated, moerland2020model}, and thereby let the global versus local distinction dominate. 

\begin{quote}
\centering \it
Planning is a class of MDP algorithms that 1) use a model and 2) only store a local solution.   
\end{quote}

\begin{quote}
\centering \it
Reinforcement learning is a class of MDP algorithms that store a global solution.
\end{quote}

\noindent The definition of reinforcement learning may then be further partitioned into model-free and model-based RL:

\begin{quote}
\centering \it
Model-free reinforcement learning is a class of MDP algorithms that 1) do not use a model, and 2) store a global solution.
\end{quote}

\begin{quote}
\centering \it
Model-based reinforcement learning is a class of MDP algorithms that 1) use a model, and 2) store a global solution.
\end{quote}

\noindent These definitions are summarized in Table \ref{tab_model_based_boundaries}. We explicitly introduce these definitions since the boundaries between both fields have generally remained vague, and a clear separation (for example between local and global solutions) will later on be useful in our framework as well. 

\begin{table}[!t]
\small
\centering
\caption{Categorization of planning and reinforcement learning, based on 1) the presence of a model (settable/reversible access to the MDP dynamics), and 2) the presence of a global/learned solution.}
\label{tab_model_based_boundaries}
\begin{tabular}{ p{6.5cm}  P{2.5cm}  P{2.5cm} }
\toprule
   & \centering \bfseries{Model} & \bfseries{Global solution} \\
\bottomrule

Planning &  + &  -  \\
Reinforcement learning & +/- &  +   \\
 
\quad Model-free reinforcement learning &  - &  +  \\
\quad Model-based reinforcement learning & + &  +    \\

\bottomrule
\end{tabular}
\end{table}

\section{Background} \label{sec_literature}
Both planning and reinforcement learning are mature research fields with a large corpus of literature. As mentioned in the Introduction, the intention of this paper is not to provide full surveys of these fields. Instead, the aim of this section is to provide a quick overview of research directions in both fields, pointing into the directions of relevant literature.

\subsection{Planning} \label{sec_planning}
{\it Planning} (or {\it search}) is a large research field within artificial intelligence \citep{russell2016artificial,lavalle2006planning}. A classic approach in MDP planning is {\it dynamic programming} (DP), of which value iteration (VI) \citep{bellman1966dynamic} and policy iteration (PI) \citep{howard1960dynamic} are classic examples. DP algorithms sweep through the entire state space, repeatedly solving small subproblems based on the Bellman optimality equation. Dynamic programming is thereby a bridging technique between planning and reinforcement learning (since it combines a model and a global representation of the solution), and would under our definitions be a form of model-based reinforcement learning. While guaranteed to converge on the optimal value function, we typically cannot store the entire solution in tabular form due to the curse of dimensionality \citep{bellman1966dynamic}. Sometimes tables may be stored more efficiently, for example through binary decision diagrams (BDD) \citep{akers1978binary,bryant1992symbolic}, or we can battle the curse of dimensionality through approximate solutions \citep{powell2007approximate,bertsekas2008approximate}, which we further discuss in the section on reinforcement learning. 

Most planning literature has focused on local solution derived from traces sampled from some start state, which are often represented as {\it trees} or {\it graphs}. Historically this starts with research on {\it uninformed search}, which studied the order of node expansion in a search tree, like {\it breadth-first search} (BFS) \citep{moore1959shortest}, {\it depth-first search} \citep{tarjan1972depth}, and {\it iterative deepening} \citep{slate1983chess}. However, most planning algorithms follow a pattern of {\it best-first search}, where we next expand the node which currently seems most promising. An early example is Dijkstra's algorithm \citep{dijkstra1959note}, which next expands the node with the current lowest path cost. Dijkstra also introduced the notions of a {\it frontier} (or open set), which is the set of states on the border of the planning tree/graph that are still candidate for expansion, and of an {\it explored states} (or closed set), which is the set of states that have already been expanded. By tracking a frontier and explored set we turn a tree search into a graph search, since it prevents the further expansion of {\it redundant} paths (multiple action sequences leading to the same state). 

We may further improve planning performance through the use of {\it heuristics} \citep{simon1958heuristic}, which in planning are often functions that provide a quick, optimistic estimate of the value of a particular state. When we apply best-first search to the sum of the path cost and admissible heuristic, we arrive at the well-known search algorithm A$^\star$ \citep{hart1968formal}, which is applicable to deterministic domains. The same approach was extended to the stochastic MDP setting as AO$^\star$ \citep{pohl1970heuristic,nilsson1971problem}. Another successful idea in the (heuristic) planning literature is the use of {\it labeling} to mark a particular state as solved (not requiring further expansion) when its value estimate is guaranteed to have converged (which happens when the state is either terminal or all of its children have been solved). Labeling can be challenging due to the potential presence of loops (which we can expand indefinitely), for which LAO$^\star$ \citep{hansen2001lao} further extends the AO$^\star$ algorithm. A survey of heuristic search is provided by \citet{pearl1984heuristics}, while \citet{kanal2012search} discuss the relation of these methods to {\it branch-and-bound} search, which has been popular in operations research.

A bridging algorithm from the planning to the learning community was {\it learning real-time} A$^\star$ (LRTA$^\star$) \citep{korf1990real}, which started to incorporate learning methodology in planning methods (and was as such one of the first model-based RL papers). This approach was later extended to the MDP setting as {Real-time Dynamic Programming} (RTDP) \citep{barto1995learning}, which performs DP updates on traces sampled from a start state distribution. {\it Labeled-RTDP} \citep{bonet2003labeled} extends RTDP through a labeling mechanism for solved states, with further improvements of RTDP provided by \citet{mcmahan2005bounded,smith2006focused,sanner2009bayesian}. 

Many planning algorithms suffer from high-memory requirements, since it is typically infeasible to store all possible states in memory. Several research lines have therefore investigated planning algorithms that have reduced memory requirements. Some well-known examples are {\it iterative deepening} depth-first search \citep{slate1983chess}, iterative deepening A$^\star$ \citep{korf1985depth}, Simplified Memory-Bounded A$^\star$ (SMA$^\star$) \citep{russell1992efficient} and recursive best-first search (RBFS) \citep{korf1993linear}. For a more extensive discussion of (heuristic) MDP planning methods we refer the reader to \citet{kolobov2012planning} and \citet{geffner2013concise}.

A different branch in planning research estimates action values based on statistical sampling techniques, better known as {\it sample-based planning}. A classic approach is {\it Monte Carlo search} (MCS) \citep{tesauro1997online}, in which we sample a number of traces for each currently available action and estimate their value as the mean return of these traces. Sample-based planning was further extended to {\it sparse sampling} \citep{kearns2002sparse}, which formed the basis for {\it Monte Carlo Tree Search} (MCTS) \citep{coulom2006efficient,kocsis2006bandit,browne2012survey}. While MCS only tracks statistics at the root of the tree search, MCTS recursively applies the same principle at deeper levels of the tree as well. Exploration and exploitation within the tree are typically based on variants of the upper confidence bounds (UCB) rule \citep{auer2002finite}. MCTS for example showed early success in the game of Go \citep{gelly2006exploration}. In the control community, there is a second branch of sample-based planning known as {\it rapidly-exploring random trees} (RRTs) \citep{lavalle1998rapidly}. In contrast to MCTS, which samples in action space to construct a tree, RRTs sample in state space and try to find an action that connects the new sampled state to the existing explicit tree in memory. 

Planning in continuous state and actions spaces, like in robotics, is typically referred to as {\it optimal control} \citep{lewis2012optimal,levine2018control}. Here, dynamics functions are often smooth and differentiable, and many algorithms therefore use a form of {\it gradient-based planning}. In this case, we directly optimize the policy for the cumulative reward objective by differentiating through the dynamics function. When the dynamics model is linear and the reward function quadratic, the solution is actually available in analytical form, better known as the linear-quadratic regulator (LQR) \citep{anderson2007optimal}. In practice, dynamics are often not linear, but this can be partly mitigated by repeatedly linearizing the dynamics around the current state (known as {iterative LQR} (iLQR) \citep{todorov2005generalized}). In the RL community, gradient-based planning is often referred to as {\it value gradients} \citep{heess2015learning}. Alternatively, we can also write the MDP problem as a non-linear programming problem (i.e., take the more black-box optimization approach), where the dynamics function for example enters as a constraint, better known as {\it direct optimal control} \citep{bock1984multiple}. Another research line treats planning as probabilistic inference \citep{botvinick2012planning,toussaint2009robot,kappen2012optimal}, where we construct message-passing algorithms to infer which actions would lead to receiving a final reward.

A popular approach in the control community is {\it model predictive control} (MPC) \citep{morari1999model}, also known as {\it receding-horizon control} \citep{mayne1990receding}, where we optimize for an action up to a certain lookahead depth, execute the best action from the plan, and then re-plan from the resulting next state (i.e., we never optimize for the full MDP horizon). Such interleaving of planning and acting \citep{mcdermott1978planning} is in the planning community often referred to as {\it decision-time} planning or {\it online} planning, where we directly need to find an action for a current state. In contrast, {\it background} or {\it offline} planning \citep{sutton2018reinforcement} uses planning operations to improve the solution for a variety of states, for example stored in a global solution.   

\subsection{Reinforcement learning} \label{sec_rl}
Reinforcement learning (RL) \citep{barto1983neuronlike,sutton2018reinforcement,wiering2012reinforcement} is a large research field within machine learning. While the planning literature is mostly organized in sub-disciplines (as discussed above), RL literature can best be covered through the range of subtopics within algorithms that have been studied. A central idea in RL is the use of {\it bootstrapping} \citep{sutton1988learning}, where we plug in a {\it learned} value estimate to improve the estimate of a state that precedes it. Literature has focused on the way we can construct these bootstrap estimates, for example distinguishing between {\it on-policy} \citep{rummery1994line} and {\it off-policy} back-ups \citep{watkins1992q}. The depth of the back-up has also received much attention in RL, where estimates of different depths can for example be combined through {\it eligibility traces} \citep{singh1996reinforcement}. We can also use multi-step methods in the off-policy setting through the use of importance sampling, where we generally reweight the back-up contribution of the next step by its probability under the optimal policy. Examples in this direction are the Tree-backup (TB($\lambda$)) algorithm \citep{precup2000eligibility} and Retrace($\lambda$) \citep{munos2016safe}.  

Reinforcement learning research has also focused on direct specification of the solution, in the form of a policy function. An important result in this direction is the {\it policy gradient theorem}  \citep{williams1992simple,sutton2000policy,sutton2018reinforcement}, which specifies an unbiased estimate of the gradient of the objective with respect to policy parameters. Policy search methods can be stabilized in various ways \citep{schulman2015trust,schulman2017proximal}, can be integrated with (gradient-based) planning \citep{deisenroth2011pilco, levine2013guided}, and have for example shown much success in robotics \citep{deisenroth2013survey}. Note that policy search can also be approached in a gradient-free way, for example through evolutionary strategies \citep{moriarty1999evolutionary,whiteson2006evolutionary}, including the successful {\it cross-entropy method} (CEM) \citep{mannor2003cross}.

A central theme in reinforcement learning research is the use of supervised learning methods to {\it approximate} the solution, which allows information to {\it generalize} between similar states (and in larger problems allow a global solution to fit in memory). Early results on function approximation include tile coding \citep{sutton1996generalization} and linear approximation \citep{bradtke1996linear}, while state-of-the-art results are achieved by the use of deep neural networks \citep{goodfellow2016deep}, whose application to RL was pioneerd by \citet{mnih2015human}. Surveys of deep reinforcement learning are provided by \citet{franccois2018introduction} and \citet{arulkumaran2017deep}.

Another fundamental theme in RL research is the balance between exploration and exploitation. Random perturbation approaches include $\epsilon$-greedy and Boltzmann exploration \citep{sutton2018reinforcement}, while other approaches, such as confidence bounds \citep{kaelbling1993learning} and Thompson sampling \citep{thompson1933likelihood}, leverage the uncertainty in an action value estimate. Another large branch in RL exploration research is {\it intrinsic motivation} \citep{chentanez2005intrinsically}, which explores based on concepts like curiosity \citep{schmidhuber1991possibility}, novelty, and model uncertainty \citep{guez2012efficient}. 

Reinforcement learning and planning have been combined in the field of model-based reinforcement learning \citep{moerland2020model,hester2012learning}. In the RL community, this idea started with {\it Dyna} \citep{sutton1990integrated}, which uses sampled data (from an irreversible environment) to learn a reversible dynamics model, and subsequently makes planning updates over this learning model to further improve the value function. Successful model-based RL algorithms include AlphaZero \citep{silver2018general}, which set superhuman performance in Go, Chess and Shogi, and Guided Policy Search \citep{levine2013guided}, which was successful in robotics tasks. We can also use a learned model for gradient-based policy updates, as for example done in PILCO \citep{deisenroth2011pilco}, while a learned backward model allows us to more quickly spread new information over the state space (known as {\it prioritized sweeping} (PS) \citep{moore1993prioritized}). A full survey of model-based reinforcement learning is provided by \citet{moerland2020model}.

Reinforcement learning research is also organized around a variety of subtopics, such as hierarchical/temporal abstraction \citep{barto2003recent}, goal setting and generalization over goals \citep{schaul2015universal}, transfer between tasks \citep{taylor2009transfer}, and multi-agent reinforcement learning \citep{busoniu2008comprehensive}. While these topics are all important, our framework solely focuses on a single agent in a single MDP optimization task. However, note that many of these topics are complementary to our framework (i.e., they could further extend it). For example, we may discover higher-level actions (hierarchical RL) to define a new, more abstract MPD, in which all of the principles of our framework are again applicable.

To summarize, this section covered some important research directions within planning and reinforcement learning. Our treatment was of course superficial, and by no means covered all relevant literature from both fields. Nevertheless, it does provide common ground on the type of literature we consider for our framework. In the next section, we will try to organize the ideas from both fields into a single framework. 

\begin{figure*}[t]
  \centering
      \includegraphics[width = 0.85\textwidth]{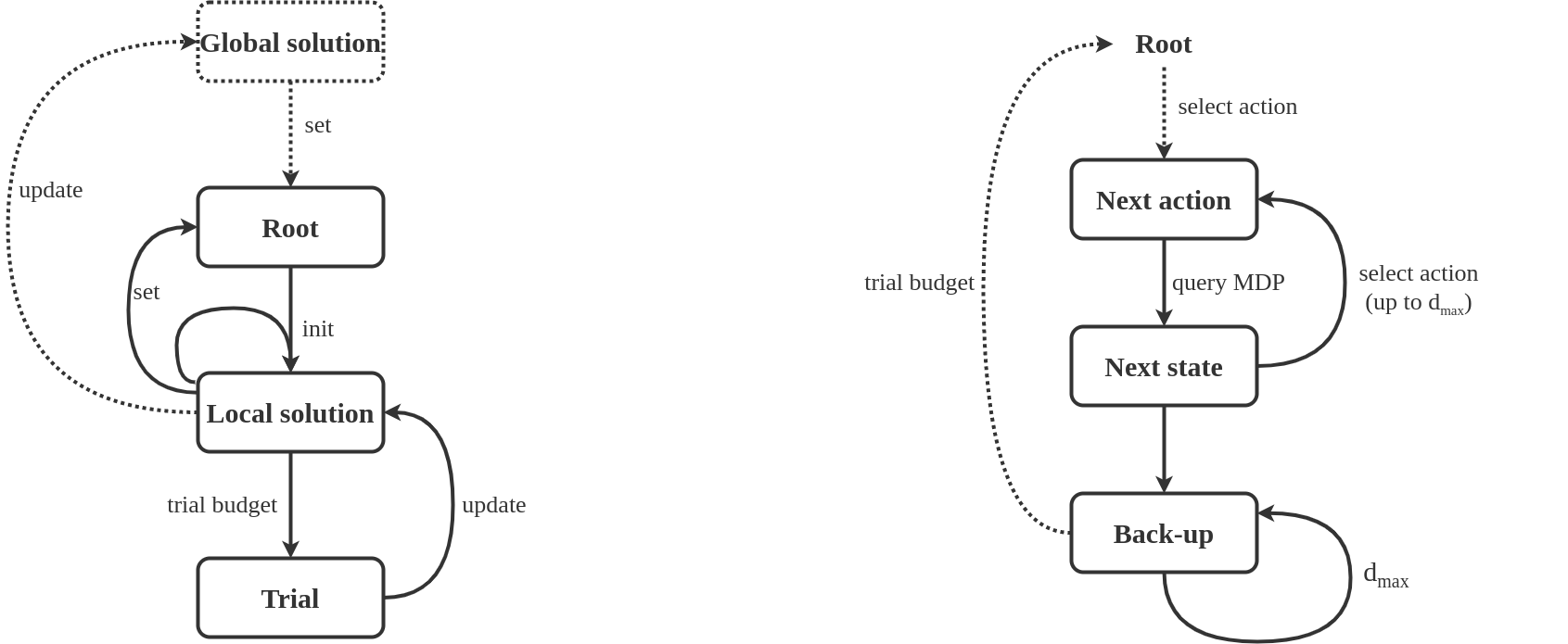}
  \caption{Graphical illustration of framework (Alg. \ref{alg_frap}). {\bf Left}: Algorithm outer loop (Alg. \ref{alg_frap}, line 4), illustrating the interplay of global and local solutions with trials. After possibly initializing a global solution, we repeatedly fix a new root state for which we want to improve our solution. Then, we initialize a new local solution for the particular root, and make one or multiple trials (trial budget), where each trial updates the local solution. After the budget is expanded, we may use the local solution to update the global solution and/or set the next root state and/or reuse information for the next local solution. The process then repeats with setting a new root, possible based on the global and/or local solution. {\bf Right}: Algorithm inner loop (Alg. \ref{alg_frap}, line 5), illustrating an individual trial. A trial starts from a root node, from which we repeatedly select actions, query the MDP at the specific state-action pair, and then transition to a next state. We repeat this process $d_{\max}$ times, after which we start the back-up phase, consisting of $d_{\max}$ back-ups. When budget is available, we start another trial from the same root node.}
    \label{fig_query_backup}
\end{figure*}

\begin{algorithm}[p]
\caption{FRAP pseudocode. In planning, there is no global solution, and the orange lines therefore disappear (and {\bf g} therefore drops from all functions as well). In model-free RL there are restrictions on the blue lines: we can only select actions and next states in a single forward trace per root, which indirectly restricts the trial budget per root (to the number of target depths we reweight over within the trace, which is often set to one) and the way we set the next root (which either has to be a next state we reached within the trial or a reset to an initial state of the MDP). In model-based RL, all elements of the framework can be active.} \label{alg_frap}
\begin{adjustwidth}{\algorithmicindent}{0pt} 
{\textbf{Input}: MDP $\mathcal{M}$, $root\textunderscore budget$ (number of root states), $trial\textunderscore budget$ (number of trials per root), $d_{max}$() (rule for maximum depth of trial).}
\end{adjustwidth}
\begin{algorithmic}[1] 
\State \textcolor{orange}{{\bf g} $\gets$ \textsc{init\textunderscore global\textunderscore solution}()} \Comment{Sec. \ref{sec_solution}} \\
$s$ $\gets$ \textsc{set\textunderscore first\textunderscore root}() \Comment{Sec. \ref{sec_set_root node}}\\
{\bf l} $\gets$ \textsc{init\textunderscore local\textunderscore solution}() \Comment{Sec. \ref{sec_solution}}\\

{\bf while} $root\textunderscore budget$ left {\bf and} not converged: \\
\qquad {\bf while} \textcolor{blue}{$trial\textunderscore budget$} left: \Comment{Sec. \ref{sec_trial_budgets}}\\
\qquad \qquad {\bf l} $\gets$ \textsc{visit\textunderscore state}($s$, {\bf l}, {\bf g}) \\

\qquad \textcolor{orange}{{\bf g} $\gets$ \textsc{update\textunderscore global\textunderscore solution}({\bf l})} \Comment{Sec. \ref{sec_update}}\\
\qquad \textcolor{blue}{$s$ $\gets$ \textsc{set\textunderscore next\textunderscore root}({\bf l}, {\bf g})} \Comment{Sec. \ref{sec_set_root node}}\\
\qquad {\bf l} $\gets$ \textsc{init\textunderscore local\textunderscore solution}(${\bf l}$) \Comment{Sec. \ref{sec_solution}} \\

\vspace{0.3cm}

\textsc{visit\textunderscore state}(s, {\bf l}, {\bf g}): \\
\qquad {\bf if} $s$ is terminal: \\
\qquad \qquad $\hat{V}(s) \gets 0$ \\
\qquad {\bf elif} $s$ at $d_{\max}$({\bf l}): \Comment{Sec. \ref{sec_trial_budgets}}\\
\qquad \qquad $\hat{V}(s) \gets $ \textsc{bootstrap}($s$, {\bf g}) \Comment{Sec. \ref{sec_bootstrap}}\\
\qquad {\bf else}: \\
\qquad \qquad $a$ $\gets$ \textsc{select\textunderscore action}($s$, {\bf l}, {\bf g}) \Comment{Sec. \ref{sec_selection}} \\ 
\qquad \qquad {\bf l } $\gets$ \textsc{visit\textunderscore action}(s, a, {\bf l}, {\bf g}) \\
\qquad \qquad $\hat{V}(s) \gets$ \textsc{backup\textunderscore policy}(s, {\bf l}, {\bf g}) \Comment{Sec. \ref{sec_backup}} \\
\qquad {\bf l} $\gets$ \textsc{update\textunderscore local\textunderscore solution}({\bf l}, $\hat{V}(s)$) \Comment{Sec. \ref{sec_update}}\\
\qquad {\bf return} {\bf l} \\

\vspace{0.3cm}

\textsc{visit\textunderscore action}(s, a, {\bf l}, {\bf g}): \\
\qquad {\bf if} $(s,a)$ at $d_{\max}$({\bf l}): \Comment{Sec. \ref{sec_trial_budgets}}\\
\qquad \qquad $\hat{Q}(s,a) \gets $ \textsc{bootstrap}($s$, a, {\bf g}) \Comment{Sec. \ref{sec_bootstrap}}\\
\qquad {\bf else}: \\
\qquad\qquad $\mathcal{T}(s'|s,a)$, $\mathcal{R}(s,a,s') \gets$ \textsc{query\textunderscore mdp}($s,a$) \Comment{Sec. \ref{sec_model_types}}\\
\qquad\qquad \textcolor{blue}{$s', r \gets$ \textsc{sample\textunderscore or\textunderscore select}($\mathcal{T}(s'|s,a)$, $\mathcal{R}(s,a,s')$)} \Comment{Sec. \ref{sec_selection}} \\ 
\qquad \qquad {\bf l} $\gets$ \textsc{visit\textunderscore state}($s'$, {\bf l}, {\bf g}) \\
\qquad\qquad $\hat{Q}(s,a) \gets$ \textsc{backup\textunderscore dynamics}($s$, $a$, $r$, {\bf l}) \Comment{Sec. \ref{sec_backup}} \\
\qquad {\bf l} $\gets$ \textsc{update\textunderscore local\textunderscore solution}({\bf l}, $\hat{Q}(s,a)$) \Comment{Sec. \ref{sec_update}}\\
\qquad {\bf return} {\bf l}
\end{algorithmic}
\end{algorithm}

\begin{sidewaystable}[p]
\centering
\small
\caption{Overview of dimensions in the Framework for Reinforcement learning and Planning (FRAP). Examples for several algorithms are shown in Table \ref{table_overview}. IM = Intrinsic Motivation.  \label{table_framework}}
\begin{tabular}{ p{4cm} p{5cm}  p{10.5cm}  }
\toprule
\bfseries{Dimension} \newline  &  \bfseries{Consideration} &  \bfseries{Choices}  \newline \\
  \hline			
 & & \\
 
1. Solution (\ref{sec_solution}) & - Coverage & Global, local \\ 
 & - Type & (Goal-conditioned) value, (goal-conditioned) policy, counts, .. \\  
 & - Method & Param. tabular, param. approximate, non/semi-parametric \\ 
 & - Initialization & Uniform, random, optimistic, expert \\ 

& & \\ 
 
2. Set root state (\ref{sec_set_root node}) & - Selection  &  Ordered, initial state, forward sampling, backward sampling, previously visited \\

& & \\

3. Budget per root (\ref{sec_trial_budgets}) & - Number of trials (width)  & 1, $n$, convergence, $\infty$ \\

 & - Depth per trial ($d_{\max}$) & 1, $n$, adaptive, $\infty$ \\

& & \\

4. Selection in trial (\ref{sec_selection}) & - Next action & Ordered, greedy (with heuristic), value-based perturbation (random, means, uncertainty), state-based perturbation (knowledge-based IM, competence-based IM) \\
 & - Next state & Sample, ordered \\
& & \\

5. Bootstrap (\ref{sec_bootstrap}) & - Location  & State, state-action \\
& - Type  & Learned, heuristic \\

& & \\

6. Back-up (\ref{sec_backup}) & - Back-up policy & Behavioral policy, greedy/max, other policy .. \\
 & - Policy expectation & Sample/partial, expected/full \\
 & - Dynamics expectation & Sample/partial, expected/full \\
 & - Additional characteristics & Explored states, convergence label, counts, uncertainty, return distribution \\

& & \\

7. Update (\ref{sec_update}) & - Loss/objective  & Squared loss, policy gradient, value gradient, cross-entropy, etc. \\ 
& - Learning rate  & Step ($\eta$ fixed), Replace ($\eta=1.0$ on table), Average ($\eta=1/n$ on table), Eligibility ($\eta = (1-\lambda) \cdot \lambda^{(d-1)} $), Adaptive (trust region), etc. \\ 
\bottomrule

\end{tabular}

\end{sidewaystable}

\section{Framework} \label{sec_unifying_view}
We will now introduce the Framework for Reinforcement Learning and Planning (FRAP). Pseudocode for the framework is provided in Algorithm \ref{alg_frap}, while all individual dimensions are summarized in Table \ref{table_framework}. We will first cover the high-level intuition of the framework, as visualized in Figure \ref{fig_query_backup}. FRAP centers around the notion of {\it root states} and {\it trials}. 

\begin{quote} \centering \it 
A root state is a state for which we attempt to improve the solution estimate. 
\end{quote}

\begin{quote} \centering \it 
A trial is a sequence of forward actions and next states from a root state, which is used to compute an estimate of the cumulative reward from the root state.
\end{quote}

The central idea of FRAP is that all planning and reinforcement learning algorithms repeatedly 1) fix root states, 2) make trials from these root states, 3) improve their solution based on the outcome of these trials, and 4) use this improved solution to better direct new trials and better set new root states. FRAP therefore consists of an {\it outer loop} (the while loop on Alg. \ref{alg_frap}, line 4), in which we repeatedly set new root states, and an {\it inner loop} (the while loop on Alg. \ref{alg_frap}, line 5), in which we (repeatedly) make trials from the current root state to update our solution. We will briefly discuss both loops. 

An schematic illustration of the outer loop is shown on the left side of Fig. \ref{fig_query_backup}. The algorithm starts by potentially initializing a global solution (for all states), and subsequently fixing a new root state. Then, we initialize a local solution for the particular root, and start making trials from the root, which each update the local solution. When we run out of trial budget for this root, we may use the local solution to update the global solution (when used). Afterwards, we fix a next root state, and initialize a new local solution, in which we may reuse information from the last local solution (Alg. \ref{alg_frap}, line 9). The outer loop then repeats for the new root state.

The inner loop of FRAP consists of trials, and is schematically visualized on the right of Fig. \ref{fig_query_backup}. A trial starts from the root node, and consists of a forward sequence of actions and resulting next states and rewards, which are obtained from {\it queries} to the MDP dynamics. This process repeats $d_{\max}$ times, where the specification of $d_{\max}$ depends on the local solution and differs between algorithms. The forward phase of the trial then halts, after which we possibly {\it bootstrap} to estimate the remaining expected return from the leaf state, without further unfolding the trial. Then, the trial proceeds with a sequence of {\it one-step back-ups}, which process the acquired information from the forward phase. We repeat the trial process until we run out of budget, after which we fix a new root state (Alg. \ref{alg_frap}, line 8). 

Action selection in FRAP not only happens within the trial (Alg. \ref{alg_frap}, line 16), but is in many algorithms also part of next root selection (Alg. \ref{alg_frap}, line 8). It is important to mention that in the case of model-free RL, where we have irreversible access to the MDP dynamics, these two action selection moments are actually equal by definition. For example, a model-free RL agent may fix a root, sample a trial from this root, and use it to update the global solution. However, because the environment is irreversible, the next root selection has to use the same action and resulting next state as was taken within the trial. Model-free RL agents therefore have some specific restrictions in the FRAP pseudocode, as illustrated on the blue lines of Alg. \ref{alg_frap} (the trial budget per root is for example also by definition equal to one). 

FRAP is therefore really a conceptual framework, and practical implementations may differ from the pseudocode in Alg. \ref{alg_frap}. For example, many planning methods store an explicit frontier, i.e., the set of nodes that are candidate for expansion. Practical implementations would directly jump to the frontier, and not first traverse the known part of the tree from the root, as happens in each trial of Alg. \ref{alg_frap}. However, it is conceptually useful to still think of these forward steps, since they will be part of the back-up phase (we are eventually looking for a good decision at the root). Another example would be a model-free RL agent that uses a Monte Carlo return estimate. Practical implementations may sample a full episode, compute the cumulative reward starting from each state in the episode, and jointly update the solution for all these states. However, conceptually every state in the episode has then been a root state once, for which we compute an estimate. In FRAP, we would therefore see this as sampling the actual episode only once from the first root, store it in the local solution, and then repeatedly set new roots along the states in the episode, where we keep reusing the local solution from the last root (Alg. \ref{alg_frap} line 9). In summary, all algorithms conceptually fit FRAP, since they all fix root states for which they compute improved estimates of the cumulative return and solution, but some algorithms may take implementation shortcuts. 

We are now ready to discuss the individual dimensions of the framework, i.e., describe the possible choices on each of the lines in Alg. \ref{alg_frap}. These dimensions are: how to {\it represent} the solution, how to {\it set the next root state}, which {\it trial budget} to allocate per root state, how to {\it select} actions and next states within a trial, how to {\it back-up} information obtained from the trial, and how to {\it update} the local and global solution based on these back-up estimates. The considerations of FRAP are summarized in Table \ref{table_framework}, while the comments on the right side of Alg. \ref{alg_frap} indicate to which lines each dimension is applicable. 

\subsection{Solution representation} \label{sec_solution}
We first of all have to decide how we will represent the solution to our problem. The top row of Table \ref{table_framework} shows the four relevant considerations: the coverage of our solution, the type of function we will represent, the method we use to represent this function, and the way we initialize the chosen method. The first item distinguishes between {\it local/partial} (for a subset of states) and {\it global} (for all states) solutions, a topic which we already extensively discussed in Sec. \ref{sec_plan_rl_definitions}. Note that FRAP {\it always} builds a local solution: even a single episode of a model-free RL algorithm is considered a local solution that estimates the value of states in the trace. A local solution therefore aggregates information from one or more trials, which may then itself be used to update a global solution (when we use one) (Alg. \ref{alg_frap}, line 1). 

\begin{table}
\small
\centering
\caption{Overview of notation. Each trial provides new back-up estimates $\hat{V}(s)$ and $\hat{Q}(s,a)$ at the states and actions that appear in the trial. These estimates are aggregated in the local solution $V^{\bf l}(s)$ and $Q^{\bf l}(s,a)$ (i.e., the local solution can be influenced by multiple trials). The local solution may itself be used to update the global solution $V^{\bf g}(s)$, $Q^{\bf g}(s,a)$ and/or $\pi^{\bf g}(a|s)$. When the global solution is stored in approximate form (which is often the case), we denote them by $V^{\bf g}_\theta(s)$, $Q^{\bf g}_\theta(s,a)$ and/or $\pi^{\bf g}_\theta(a|s)$ (where $\theta$ denotes the parameters of the approximation). Back-up estimates and local solutions are in practice never represented in approximate form.} \label{table_solution_types}
\begin{center}
\begin{tabular}{l c c c }
& {\bf Back-up estimate} & {\bf Local solution} & {\bf Global solution} \\
\toprule
{\bf Tabular} & $\hat{V}(s)$, $\hat{Q}(s,a)$ &  $V^{\bf l}(s)$, $Q^{\bf l}(s,a)$ & $V^{\bf g}(s)$, $Q^{\bf g}(s,a)$, $\pi^{\bf g}(a|s)$ \\
{\bf Approximate} & (-) & (-) & $V^{\bf g}_\theta(s)$, $Q^{\bf g}_\theta(s,a)$, $\pi^{\bf g}_\theta(a|s)$  \\
\bottomrule
\end{tabular}
\end{center}
\end{table}

For both local and global solutions we next need to decide what type of function to represent. The most common choices are to represent the solution as a {\it value} function $V: \mathcal{S} \to \mathbb{R}$, {\it state-action value} function $Q: \mathcal{S} \times \mathcal{A} \to \mathbb{R}$, or {\it policy} function $\pi: \mathcal{S} \to p(\mathcal{A})$. Some algorithms combine value and policy solutions, better known as {\it actor-critic} algorithms \citep{konda1999actor}. We may also store the {\it uncertainty} around value estimates \citep{osband2016deep,moerland2017efficient}, for example using {\it counts} \citep{kocsis2006bandit}, or through convergence labels that mark a particular value estimate as solved \citep{nilsson1971problem,bonet2003labeled}. Some methods also store the entire distribution of returns \citep{bellemare2017distributional,moerland2018potential}, or condition their solution on a particular goal \citep{schaul2015universal} (i.e., store a solution for multiple reward functions).  

After deciding on the type of function to represent, we next need to specify the representation method. This is actually a supervised learning question, which we can largely break up in {\it parametric} and {\it non-parametric} approaches. {\it Parametric tabular} representations use a unique parameter for the solution at each state-action pair, which is for example used in the local solution of a graph search, or in the global solution of a tabular RL algorithm. For high-dimensional problems, we typically need to use {\it parametric approximate} representations, such as (deep) neural networks \citep{rumelhart1986learning,goodfellow2016deep}. Apart from reduced memory requirement, a major benefit of approximate representations it their ability to {\it generalize} over the input space, and thereby make predictions for state-actions that have not been observed yet. However, the individual predictions of approximate methods may contain errors, and there are indications that the combination of tabular and approximate representations may provide the best of both worlds \citep{silver2017mastering,langlois2019benchmarking,moerland2020think}. Alternatively, we may also store the solution in a {\it non-parametric} way, where we simply store exact sampled traces (e.g., a search tree that does not aggregate over different traces), or {\it semi-parametric} way \citep{graves2016hybrid}, where we may optimize a neural network to write to and read to a table \citep{blundell2016model,pritzel2017neural}, sometimes referred to as {\it episodic memory} \citep{gershman2017reinforcement}. 

Finally, we also need to initialize our solution representation. Tabular representations are often {\it uniformly} initialized, for example setting all initial estimates to 0. Approximate representations are often {\it randomly} initialized, which provides the tie breaking necessary for gradient-based updating. Some approaches use initialization to guide exploration, either through {\it optimistic initialization} (when a state has not been visited yet, we consider its value estimate to be high) \citep{bertsekas1996neuro} or {\it expert initialization} (where we use imitation learning from (human) expert demonstrations to initialize the solution) \citep{hussein2017imitation}. We will further discuss exploration methods in Sec. \ref{sec_selection}. 

An overview of our notation for the different local/global and tabular/approximate solution types is shown in Table \ref{table_solution_types}. We will denote {\it local} estimates with superscript ${\bf l}$, e.g., $V^{\bf l}(s)$ or $Q^{\bf l}(s,a)$, and {\it global} solutions with superscript {\bf g}, e.g., $V^{\bf g}(s)$, $Q^{\bf g}(s,a)$ or $\pi^{\bf g}(a|s)$. In practice, only global solutions are learned in approximate form, which we indicate with a subscript $\theta$ (for parameters $\theta$). 

As you will notice, Table \ref{table_solution_types} contains a separate entry for the {\it back-up estimate}, $\hat{V}(s)$ or $\hat{Q}(s,a)$, which are formed during every trial. Especially researchers from a planning background may find this confusing, since in many algorithms the back-up estimate and local solution are actually the same. However, we should consider these two different quantities, for two reasons. First of all, in some algorithms, like the roll-out phase of MCTS, we do make additional MDP queries (the trial continues) and back-ups, but the back-up estimate from the last part of the trial is never stored in the local solution (the local solution expands with only one new node per trial). Second, many algorithms use their local solution to {\it aggregate} cumulative reward estimates from different depths, which is for example used in eligibility traces \citep{sutton2018reinforcement}. For our conceptual framework, we therefore consider each cumulative reward estimate the result of a single trial, and the local solution may combine the estimate of trials in multiple ways. We will discuss ways to aggregate back-up estimates into the local solution in Sec. \ref{sec_update}. 

\begin{figure}[!t]
  \centering
      \includegraphics[width = 0.6\textwidth]{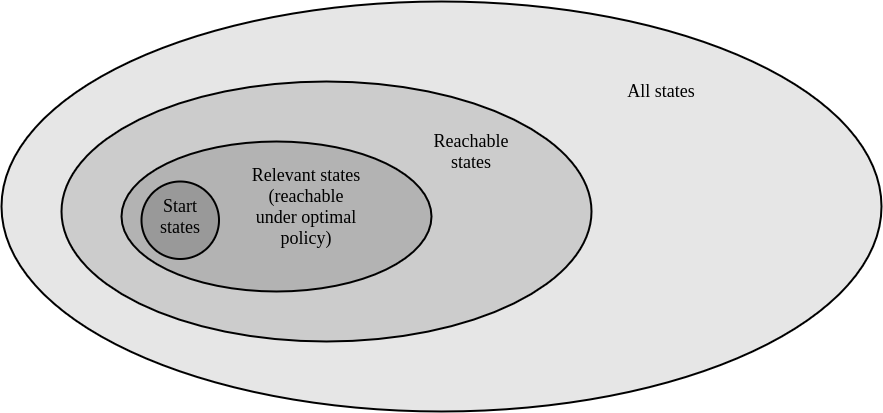}
  \caption{Venn diagram of total state space. Only a subset of the entire state space is {\it reachable} from the start state under {\it any policy}. An even smaller subset of the reachable set is eventually {\it relevant}, in the sense that they are reachable from the start state under the {\it optimal policy}. Finally, a subset of the relevant state are of course all start states. Figure extended from \citet{sutton2018reinforcement}.}
    \label{fig_relevant_states}
\end{figure}

\subsection{Set a root state} \label{sec_set_root node}
The next consideration in our framework is the selection of a root state (Alg. \ref{alg_frap}, line 2 \& 8), for which we will attempt to improve our solution (by computing a new value estimate). The main considerations are listed in the second row of Table \ref{table_framework}. A first approach is to select a state from the state space in an {\it ordered} way, for a example by sweeping through all possible states \citep{bellman1966dynamic,howard1960dynamic}. A major downside of this approach is that many states in the state space are often not even reachable from the start state (Fig. \ref{fig_relevant_states}), and we may spend much computational effort on states that will never be part of the practical solution. 

When the MDP definition includes the notion of a {\it start state distribution}, this information may be utilized to improve our selection of root states, by only sampling root states on traces from the start. This ensures that new roots are always reachable, which may strongly reduce the number of states we will update in practice (illustrated in Fig. \ref{fig_relevant_states}). In Table \ref{table_framework}, we list this as the {\it forward sampling} approach to selecting new root states. Note that this generally also involves an action selection question (in which direction do we set the next root), which we will discuss in Sec. \ref{sec_selection}. 

The next option is to select new root states in the reverse direction, i.e., through backward sampling (instead of forward sampling). This approach does require a {\it backwards model} $p(s,a|s')$, which specifies the possible state-action pairs $(s,a)$ that may lead to a next state $s'$. The main idea is to set next root states at the possible precursor states of a state whose value has just changed much, better known as {\it prioritized sweeping} \citep{moore1993prioritized}. We thereby focus our update budget on regions of the state space that likely need updating, which may speed-up convergence. Similar ideas have been studied in the planning community as {\it backward search} or {\it regression search} \citep{nilsson1982principles,bonet2001planning,alcazar2013revisiting}, which makes prioritized sweeping an interleaved form of forward and backward search.

Finally, we do not always need to select the next root state from the current trace. For example, we may track the set of {\it previously visited states}, and select our next root from this set. This approach, which is for example part of Dyna \citep{sutton1990integrated}, gives greater freedom in the order of root states, while it still ensures that we only update reachable states. To summarize, we need to decide on a way to set root states, which may for example be done in an ordered way, through forward sampling, through backward sampling, or by selecting previously visited states (Table \ref{table_framework}, second row). 


\subsection{Budget per root} \label{sec_trial_budgets}
\begin{figure}[!t]
  \centering
      \includegraphics[width = 0.7\textwidth]{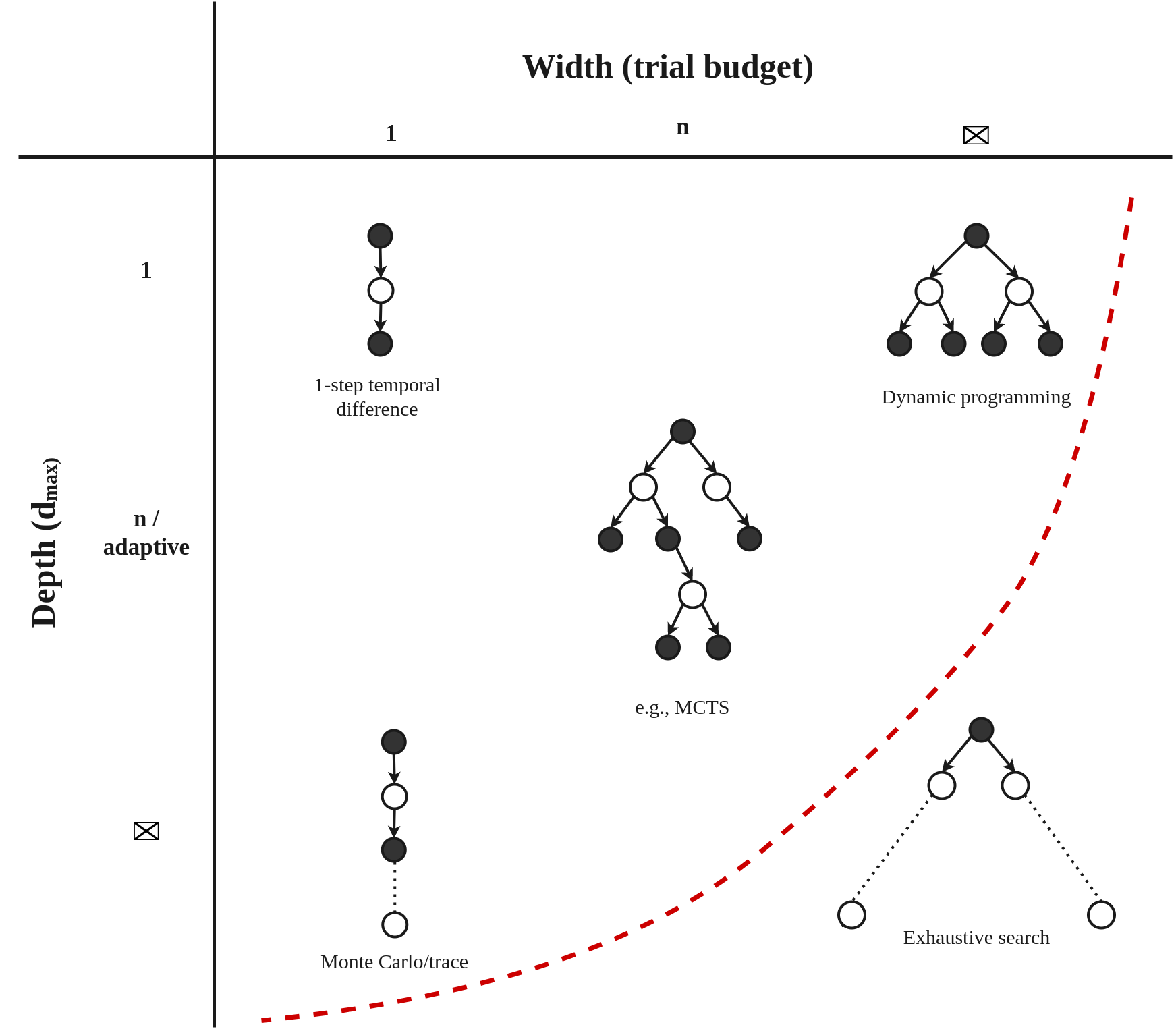}
  \caption{Possible combinations of width (trial budget) and depth ($d_{\max}$) per trial from a root state. Practical algorithms reside somewhere left of the left dotted line, since full with combined with full depth (exhaustive search) is not feasible in larger problems. Figure extended from \citet{sutton2018reinforcement}.}
    \label{fig_width_depth_frap}
\end{figure}

After we fixed a root state (a state for which we will attempt to improve the solution), we need to decide on 1) the number of trials from the particular root (Alg. \ref{alg_frap} line 5), and 2) when a trial itself will end, i.e., the depth $d_{\max}$ of each forward trial (Alg. \ref{alg_frap} line 13 \& 22). These possible choices on each of these two considerations are listed in the third row of Table \ref{table_framework}. Note that since every trial consists of a single forward beam, the total number of trials is actually a good measure of the total width of the local solution (Fig. \ref{fig_trial_illustration}). The joint space of both considerations is visualized in Fig. \ref{fig_width_depth_frap}, which we will discuss below.

Regarding the {\it trial budget per root state}, a first possible choice is to only run a single trial. This choice is characteristic for model-free RL algorithms \citep{sutton2018reinforcement}. Algorithms that have access to a model may also run multiple trials per root state. This budget can for example be specified as a fixed hyperparameter, as is often the choice in MCTS \citep{browne2012survey}. When we interact with a real-world environment, the trial budget may actually be enforced by the time until the next decision is required. In the planning community, this is referred to as {\it decision time planning} or {\it online planning}. In offline approaches, we may also provide an adaptive trial budget, for example until some convergence criterion is met (often in combination with an admissible heuristic, which may reduce the required number of trials to convergence a lot) \citep{nilsson1971problem,hansen2001lao,bonet2003labeled}. Finally, we may also specify an infinite trial budget, i.e., we will repeat trials until all possible sequences (for the specified depth) have been expanded. 

The second decision involves the {\it depth} of each individual trial. A first option is to use a trial depth of one, which is for example part of value/policy iteration \citep{bellman1966dynamic} and temporal difference learning \citep{sutton1988learning,watkins1992q,rummery1994line}. We may also specify a fixed multi-step depth, which is the case for $n$-step methods, or specify a full depth ($\infty$), in which case we unroll the trail until a terminal state is reached (in practice we often still limit the trial by a large depth). The latter is also known as a {\it Monte Carlo roll-out}, which is for example used in MCTS. Finally, many algorithms make use of an {\it adaptive} trial depth, which depends on the current local solution (i.e., note that $d_{\max}({\bf l})$ depends on {\bf l} in Alg. \ref{alg_frap}, lines 13 and 22). For example, several (heuristic) planning algorithms terminate a trial once we reach a state or action that did not appear in our current local solution yet \citep{hart1968formal,nilsson1971problem}. As another example, we may terminate a trial once it reaches a state in the explored set or makes a cycle to a duplicate state, which are also examples of an adaptive $d_{\max}({\bf l})$. To summarize, the trial budget and depth of each trial are important considerations in all planning and RL algorithms. 

\subsection{Selection within a trial} \label{sec_selection}
Once we have specified the trial budget and depth rules from a particular root state, we have to decide how to actually select the actions and states that will appear in each individual trial (they may unroll in different directions). In other words, we have specified the overall shape of all trials in Fig. \ref{fig_width_depth_frap}, but not yet how this shape will actually be unfolded. We will first discuss {\it action selection}, which happens in Alg. \ref{alg_frap} line 16 and in many algorithms also at line 8, when we set the next root through forward sampling. Afterwards, we will discuss {\it next state selection}, which happens in line 26 of Alg. \ref{alg_frap}. The considerations that we discuss for both topics are listed in the fourth row of Table \ref{table_framework}.

\paragraph{Action selection} 
The first approach to action selection is to pick actions in an {\it ordered} way, where we select actions {\it independently} of our interaction history with the MDP. Examples include uninformed search methods, such as iterative. A downside of ordered action selection is that it may spend much time on states with lower value estimates, which typically makes it infeasible in larger problems. Most methods therefore try to prioritize actions in trials based on knowledge from previous trials. A first category of approaches prioritize actions based on their (current) value estimate, which we will call {\it value-based selection}. The cardinal example of value-based selection is {\it greedy} action selection, which repeatedly selects actions with the highest current value estimate. This is the dominant approach in the heuristic search literature \citep{hart1968formal,nilsson1971problem,hansen2001lao,barto1995learning}, where an {\it admissible} heuristic may guarantee that greedy action selection will find the optimal solution.

\begin{figure*}[t]
  \centering
      \includegraphics[width = 1.0\textwidth]{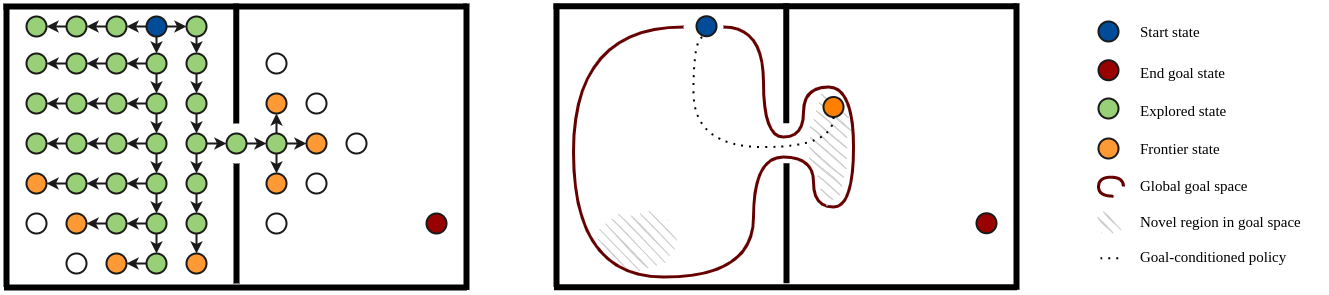}
  \caption{Frontier-based exploration in planning (left) and reinforcement learning (right, {\it intrinsic motivation}). {\bf Left}: Frontier and explored set in a graph. Blue denotes the start state, red a final state, green denotes the explored set (states that have been visited and whose successors have been visited), orange denotes the frontier (states that have been visited but whose successors have not all been visited). Methods without a frontier and explored set (like random perturbation, which is used in most RL approaches) may sample many redundant trials that make loops in the left part of the problem, because they do not find the narrow passage. {\bf Right}: In large problems, it may become infeasible to store the frontier and explored set in tabular form. Part of intrinsic motivation literature \citep{colas2020intrinsically} tracks {\it global} (sub)goal spaces (red line) in global, approximate form. We may for example sample new goals from this space based on novelty, and subsequently attempt to reach that goal through a goal-conditioned policy, effectively mimicking frontier-based exploration in approximate, global form.}
    \label{fig_frontier}
\end{figure*}

Note that heuristic search algorithms in practice usually maintain a {\it frontier} (Fig. \ref{fig_frontier}), and therefore do not actually need to greedily traverse the local solution towards the best leaf state. However, as \citet{schulte2014balancing} also show, any ordering on the frontier can also be achieved by step-wise action selection from the root, and frontiers therefore conceptually fully fit into our framework (although the practical implementation may differ). The notion of frontiers is important, because algorithms that use a frontier often {\it switch} their action selection strategy once they reach the frontier. For example, a heuristic search algorithm may greedily select actions within the known part of the local solution, but at the frontier expand all possible actions, which is a form of ordered action selection. For some algorithms, we will therefore separately mention the action selection strategy {\it before the frontier} (BF) and {\it after the frontier} (AF).

Without an admissible heuristic greedy action selection is not guaranteed to find the optimal solution. Algorithms therefore usually introduce a form of {\it exploration}. A first option in this category is {\it random perturbation}, which is in the RL community usually referred to as $\epsilon$-greedy exploration \citep{sutton2018reinforcement}. Similar ideas have been extensively studied in the planning community \citep{valenzano2014comparison}, for example in limited discrepancy search \citep{harvey1995limited}, $k$-best-first-search (KBFS) \citep{felner2003kbfs} and best-first width search (BFWS) \citep{lipovetzky2017best}. We may also make the selection probabilities proportional to the current mean estimates of each action, which is for discrete and continuous action spaces for example achieved by Boltzmann exploration \citep{cesa2017boltzmann} and entropy regularization \citep{peters2010relative}.

\begin{table}[t]
\centering
\small
\caption{Overview of action selection methodology within a trial. At the highest level, we may either prioritized actions in an ordered way (independent of our interaction history with the MDP), in a value-based way (based on obtained rewards in our interaction history with the MDP), or in astate-based (based on our interaction history with the MDP, but independent of the value). The table shows possible subcategories, and some characteristic examples in the right column. \label{table_action_selection}}
\begin{tabular}{ p{6.0cm}  p{8.0cm} }

{\bf Action selection method} & {\bf Characteristic examples} \\

  \toprule	

{\bf Ordered} & Value iteration \citep{bellman1966dynamic} \newline Iterative deepening \citep{korf1985depth} \\

&   \\
{\bf Value-based} & \\
\quad - Greedy (with heuristic) & AO$^\star$ \citep{nilsson1971problem} \newline RTDP \citep{barto1995learning}  \\
\quad - Random perturbation & $\epsilon$-greedy \citep{sutton2018reinforcement} \newline Gaussian noise \citep{van2007reinforcement} \\
\quad - Mean perturbation & Boltzmann \citep{cesa2017boltzmann} \newline Entropy regularization \citep{peters2010relative} \\
\quad - Uncertainty perturbation & Upper confidence bounds \citep{kaelbling1993learning} \newline Posterior sampling \citep{thompson1933likelihood} \\

&   \\

{\bf State-based} & \\
\quad - Knowledge-based IM & Novelty \citep{brafman2002r} \newline Suprise \citep{achiam2017surprise} \\
\quad - Competence-based IM & Learning progress \citep{pere2018unsupervised} \newline 
Goal-reaching success \citep{florensa2018automatic} \\

\bottomrule
\end{tabular}
\end{table}

A downside of random perturbation methods is their inability to naturally transition from exploration to exploitation. A solution is to track the uncertainty of value estimate of each action, i.e., {\it uncertainty-based perturbation}. Such approaches have been extensively studied in the multi-armed bandit literature \citep{slivkins2019introduction}, and successful exploration methods from RL and planning \citep{kocsis2006bandit,kaelbling1993learning,hao2019bootstrapping} are actually based on work from the bandit literature \citep{auer2002finite}. Note that uncertainty estimation in sequential problems, like the MDP formulation, is harder than the multi-armed bandit setting, since we need to take the uncertainty in the value estimates of future states into account \citep{dearden1998bayesian,moerland2017efficient}. As an alternative, we may also estimate uncertainty in a Bayesian way, and for example explore through Thompson sampling \citep{thompson1933likelihood,osband2016deep}. Note that {\it optimistic initialization} of the solution, already discussed in Sec. \ref{sec_solution}, also uses optimism in the face of uncertainty to guide exploration, although it does not track the true uncertainty in the value estimates. 

In contrast to value-based perturbation, we may also use {\it state-based perturbation}, where we inject exploration noise {\it based on our interaction history with the MDP} (i.e., independently of the extrinsic reward). As a classic example, a particular state might be interesting because it is novel, i.e., we have not visited it before in our current interaction history with the MDP. In the reinforcement learning literature, this approach is often referred to as {\it intrinsic motivation} (IM) \citep{chentanez2005intrinsically,oudeyer2007intrinsic}. We already encountered the same idea in the planning literature through the use of frontiers and explored set, which essentially prevent expansion of a state that we already visited before. In RL (intrinsic motivation) literature, we usually make a separation between {\it knowledge-based} intrinsic motivation, which marks states or actions as interesting because they provide new knowledge about the MDP, and {\it competence-based} intrinsic motivation, where we prioritize target states based on our {\it ability} to reach them. Examples of the knowledge-based IM include intrinsic rewards for {\it novelty} \citep{brafman2002r,bellemare2016unifying}, recency \citep{sutton1990integrated}, curiosity \citep{pathak2017curiosity}, surprise \citep{achiam2017surprise}, and model uncertainty \citep{houthooft2016vime}, while we may also provide intrinsic motivation for the {\it content} of a state, for example a saliency for objects \citet{kulkarni2016hierarchical}. Competence-based IM may for example prioritize (goal) states of intermediate difficulty (which we manage to reach sometimes) \citep{florensa2018automatic}, or on which we are currently making learning progress \citep{baranes2013active,matiisen2017teacher,lopes2012exploration}. 

As mentioned above, there is clear connection between the use of frontiers in planning literature and the use of intrinsic motivation in reinforcement learning literature, which we illustrate in Fig. \ref{fig_frontier}. On the one hand, the planning literature has many techniques to track and prioritize frontiers, but these tabular approaches do suffer in high-dimensional problems. In contrast, in RL methods that do not track frontiers (but for example use random perturbation) many trials may not hit a new state at all \citep{ecoffet2021first}. Intrinsic motivation literature has studied the use of {\it global, approximate frontiers} (i.e., global, approximate sets of interesting states to explore), which is typically referred to as intrinsically motivated goal exploration processes (IMGEP) \citep{colas2020intrinsically}. An successful example algorithm in this class is Go-Explore \citep{ecoffet2021first}, which achieved state-of-the-art performance on the sparse-reward benchmark task Montezuma's Revenge. However, IMGEP approaches have their challenges as well, especially because it is hard to track convergence of approximate solutions, and our goal space may for example be off, or we do encounter a novel region but after an update of our goal-conditioned policy we are not able to get back. Tabular solutions from the planning literature do not suffer from these issues, and we conjecture that there is much potential here in the combination of ideas from both research fields. 

As mentioned in the beginning, action selection often also plays a role on Alg. \ref{alg_frap} line 8, when we select next root states through forward sampling from the previous root (as discussed in Sec. \ref{sec_set_root node}). In the planning literature, this is often referred to as the {\it recommendation function} \citep{keller2013trial} (what action do we recommend at the root after all trials and back-ups). When we want to maximize performance, action recommendation is often greedy, for example based on the visitation counts at the root of an MCTS search \citep{browne2012survey}. However, during offline learning, we may inject additional exploration into action selection at the root, for example by {\it planning to explore} (the trials in a learned model direct the agent towards interesting new root state in the true environment) \citep{sekar2020planning}. We will refer to this type of action selection as {\it next root} (NR) selection, and note that some algorithms therefore have three different action selection strategies: before the frontier (BF) within a trial, after the frontier (AF) within a trial, and to set the next root (NR) for new trials. An overview of the discussed action selection methods, with some characteristic examples, is provided in Table \ref{table_action_selection}. 

\begin{figure}[!t]
  \centering
      \includegraphics[width = 0.7\textwidth]{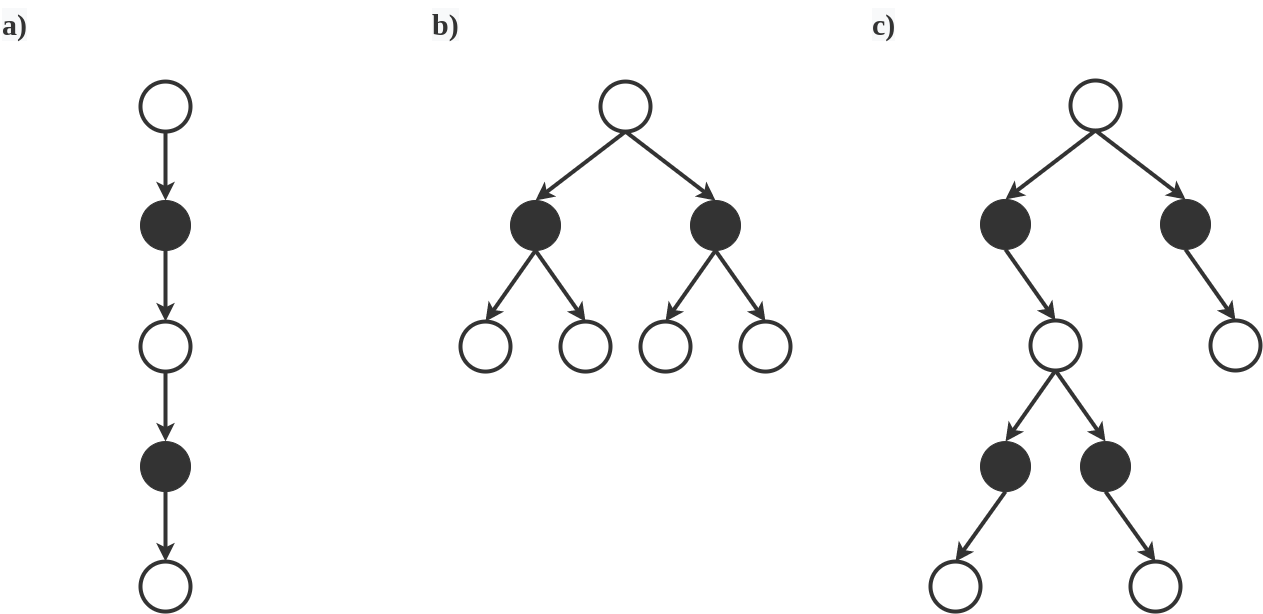}
  \caption{Example local solution patterns. {\bf a}) Local solution consisting of a single trial with depth 2. Total queries to the MDP = 2. Example: two-step temporal difference learning. {\bf b}) Local solution consisting of four trial with depth 1. Total queries to the MDP = 4. Example: value iteration. {\bf c}) Local solution consisting of three trials, one with depth 1 and two with depth 2. Total queries to the MDP = 4. Example: Monte Carlo Tree Search.}
    \label{fig_trial_illustration}
\end{figure} 

\paragraph{State selection} After our extensive discussion of action selection methods within a trial, we also need to discuss {\it next state selection}, which happens at line 26 of Alg. \ref{alg_frap}. The two possible options here are ordered and sample selection. {\it Ordered} next state selection is for example used in value and policy iteration, where we simply expand every possible next state of an action. This approach is only feasible when we have settable, descriptive access to the MDP dynamics (see Sec. \ref{sec_model_types}), since we can then decide ourselves which next state we want to make our next MDP query from. The second option is to {\it sample} the next action, which is by definition the choice when we only have generative access to the MDP dynamics. However, sampled next state selection may even be beneficial when we do have descriptive access \citep{sutton2018reinforcement}.
 
To summarize this section on action and next state selection within a trial, Figure \ref{fig_trial_illustration} illustrates some characteristic trial patterns. On the left of the figure we visualize a local solution consisting of a single trial with $d_{\max}=2$, which is for example used in two-step temporal difference (TD) learning \citep{sutton1988learning}. In the middle, we see a local solution consisting of four trials, each with a $d_{\max}$ of 1. Each action and next state is selected in an ordered way, which is for example used in value iteration \citep{bellman1966dynamic}. Finally, the right side of the figure shows a local solution consisting of three trials, one with $d_{\max}=1$ and two with $d_{\max}=2$, which could for example appear in Monte Carlo Tree Search \citep{kocsis2006bandit}. With the methodology described in this section, we can construct any other preferred local solution pattern. In the next section we will discuss what to do at the leaf states of these patterns, i.e., what to do when we reach the trial's $d_{\max}$. 

\subsection{Bootstrap} \label{sec_bootstrap}
The main aim of trials is to provide a new/improved estimate of the value of each action at the root, i.e., the expected cumulative sum of rewards from this state-action (Eq. \ref{eq_cum_reward}). However, when we choose to end a trial before we can evaluate the entire sum, we may still obtain an estimate of the cumulative reward through {\it bootstrapping}. A bootstrap function is a function that provides a quick estimate of the value of a particular state or state-action. When we decide to end our trial at a state, we need to bootstrap a state value (Alg. \ref{alg_frap}, line 14), and when we decide to end the trial at an action, we need to bootstrap a state-action value (Alg. \ref{alg_frap}, line 23). A potential benefit of a state value function is that it has lower dimension and might be easier to learn/obtain, while a state-action value function has the benefit that it allows for off-policy back-ups (see Sec. \ref{sec_backup}) without additional queries to the MDP. Note that terminal states have a value of 0 by definition. 

The bootstrap function itself may either be obtained from a {\it heuristic function}, or it can be learned. Heuristic functions have been studied extensively in the planning community. A heuristic is called {\it admissible} when it provides an {\it optimistic} estimate of the remaining value for every state, which allows for greedy action selection strategies during the search. Heuristics can be obtained from prior knowledge, but much research has focused on automatic ways to obtain heuristics, often by first solving a simplified version of the problem. When the problem is stochastic, a popular approach is {\it determinization}, where we first solve a deterministic version of the MDP to obtain a heuristic for the full planning task \citep{hoffmann2001ff,yoon2007ff}, or {\it delete relaxations} \citep{bonet2001planning}, where we temporarily ignore the action effects that remove state attributes (which is only applicable in symbolic states spaces). A heuristic is called 'blind' when it is initialized to the same value everywhere. For an extensive discussion of ways to obtain heuristics we refer the reader to \citet{pearl1984heuristics,edelkamp2011heuristic}.

The alternative approach is to {\it learn} a global state or state-action value function. Note that this function can also serve as our solution representation (see Sec. \ref{sec_solution}). The learned value function can be trained on the root value estimates of previous trials (see Sec. \ref{sec_update}), and thereby gradually improve its performance \citep{sutton1988learning,korf1990real}. A major benefit of learned value functions is 1) their ability to improve performance with more data, and 2) their ability to {\it generalize} when learned in approximate form. For example, while Deep Blue \citep{campbell2002deep}, the first computer programme to defeat a human Chess world champion, used a heuristic bootstrap function, this approach was later outperformed by AlphaZero \citep{silver2018general}, which uses a deep neural network to learn a bootstrap function that provides better generalization.

\begin{figure}[t]
  \centering
      \includegraphics[width = 1.0\textwidth]{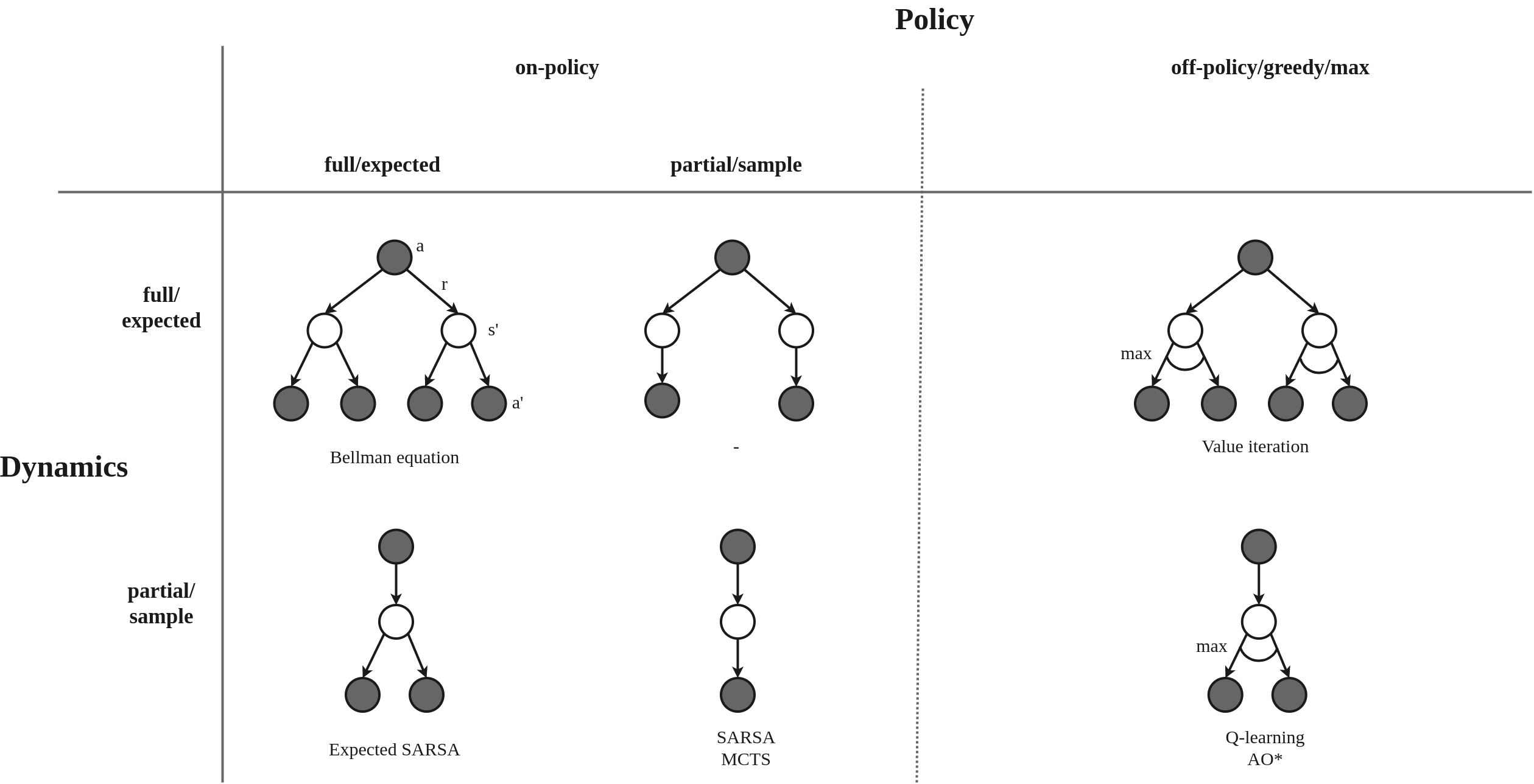}
  \caption{Types of 1-step back-ups. For the back-up over the policy (columns), we need to decide on i) the type of policy (on-policy or off-policy) and ii) whether we do a full or partial back-up. For the back-up over the dynamics (rows), we also need to decide whether we do a full or partial back-up. Note that for the greedy/max back-up policy the expected and sample back-ups are equivalent. Mentioned algorithms: Value Iteration \citep{bellman1966dynamic}, Expected SARSA \citep{van2009theoretical}, SARSA \citep{rummery1994line}, MCTS \citep{kocsis2006bandit}, Q-learning \citep{watkins1992q}, and AO$^\star$ \citep{nilsson1971problem}.} 
    \label{fig_backup_types}
\end{figure}

\subsection{Back-up} \label{sec_backup}
Bootstrapping ends the forward phase of a trial, after which we start the back-up phase (Fig. \ref{fig_query_backup}, right). The goal of back-ups is to process the acquired information of the trial. We will primarily focus on the {\it value back-up}, where we construct new estimates $\hat{V}(s)$ and $\hat{Q}(s,a)$ for states and actions that appear in the trial. At the end of this section, we will also briefly comment on other types of information we may include in the back-up. 

Value back-ups are based on the one-step Bellman equation, as shown in Eq. \ref{eq_bellman}). The first expectation of this equation, over the possible next states, shows the {\it dynamics back-up}: we need to aggregate value estimates for different possible next states into an state-action value estimate for the state-action that may lead to them. The second expectation, over the possible actions, shows the {\it policy back-up}: we want to aggregate state-action values into a value estimate at the particular state. We therefore need to discuss how to deal with width (expectations) over the policy and dynamics. In Alg. \ref{alg_frap}, policy and dynamics back-ups happen at line 18 and 28, while we will now discuss the relevant considerations for these back-ups, as listed in the sixth row of Table \ref{table_framework}.

For the policy back-up, we first need to specify which back-up policy we will actually employ. A first option is to use the current behavioural policy (which we used for action selection within the trial) as the back-up policy, which is in RL literature usually referred to as {\it on-policy} back-ups. An alternative is to use another policy than the behavioural policy, which is referred to as {\it off-policy}. The most common off-policy back-up is the {\it greedy} or {\it max} back-up, which puts all probability on the action with the highest current value estimate. The greedy back-up is common in tabular solutions, but can be unstable when combined with a global approximate solutions and bootstrapping \citep{van2018deep}. Note that off-policy back-ups do not need to be greedy, and we may also use back-up policies that are more greedy than the exploration policy, but less greedy than the max operator \citep{keller2015anytime,coulom2006efficient}.

We next need to decide whether we will make a {\it full} / {\it expected} policy back-up, or a {\it partial} / {\it sample} policy back-up. Expected back-ups evaluate the full expectation over the policy probabilities, and therefore need to expand all child actions of a state. In contrast, sample back-ups only back-up the value from a sampled action, and therefore do not need to trial all child actions (and are therefore called `partial'). Sample back-ups are less accurate but computationally cheaper, and will move towards the true value over multiple samples.

The same consideration actually applies to the back-up over the dynamics, which can also be {\it full} / {\it expected} back-up, or {\it partial} / {\it sample}. Which type of dynamics back-up we can make also depends on the type of access we have to the MDP. When we only have generative access to the MDP, we are forced to make sample back-ups. In contrast, when we have descriptive access to the MDP, we can either make expected or sample back-ups. Although sample back-ups have higher variance, they are computationally cheaper and may be more efficient when many next states have a small probability \citep{sutton2018reinforcement}. We summarize the common back-up equations for policy and dynamics in Table \ref{tab_backup_equations}, while Figure \ref{fig_backup_types} visualizes common combinations of these as back-up diagrams.

\begin{table}
\caption{Equations for the policy and dynamics back-up, applicable to Alg. \ref{alg_frap} line 18 and 28, respectively.} \label{tab_backup_equations}
\centering \small 
\begin{tabular}{p{3cm} p{6.5cm} p{2.3cm}}
 & {\bf Equation} \\
 \toprule
{\bf Policy} & \\ 
\quad Sample back-up & $ \hat{V}(s)  \gets \hat{Q}(s,a)$, & for $a \sim \pi(\cdot|s')$ \\
\quad Expected back-up & $ \hat{V}(s)  \gets \mathbb{E}_{a \sim \pi(\cdot|s)} [ \hat{Q}(s,a) ]$ & \\
\quad Greedy back-up & $ \hat{V}(s)  \gets \max_{a} [ \hat{Q}(s,a) ]$ & \\

{\bf Dynamics} & \\

\quad Sample back-up & $ \hat{Q}(s,a)  \gets \mathcal{R}(s,a,s') + \gamma \cdot \hat{V}(s')$, & for  $s' \sim T(\cdot|s,a)$ \\
\quad Expected back-up & $ \hat{Q}(s,a)  \gets \mathbb{E}_{s' \sim \mathcal{T}(s'|s,a)} [ \mathcal{R}(s,a,s') + \gamma \cdot \hat{V}(s')]$ & \\
\end{tabular}
\end{table}

Many algorithms back-up additional information to improve action selection in future trials. We may want to track the uncertainty in the value estimates, for example by backing-up visitation counts \citep{browne2012survey}, by backing-up entire uncertainty distributions around value estimates \citep{dearden1998bayesian,deisenroth2011pilco}, or by backing-up the distribution of the return \citep{bellemare2017distributional}. Some methods back-up {\it labels} that mark a particular value estimate as `solved' when we are completely certain about its value estimate \citep{nilsson1971problem,bonet2003labeled}. As mentioned before, graph searches also back-up information about frontiers and explored sets, which can be seen as another kind of label, one that removes duplicates and marks expanded states. The overarching theme in all these additional back-ups is that they track some kind of uncertainty about the value of a particular state, which can be utilized during action selection in future trials. 

\subsection{Update} \label{sec_update}
The last step of the framework involves updating the local solutions ($V^{\bf l}(s)$ and $Q^{\bf l}(s,a)$) based on the back-up estimates ($\hat{V}(s)$ and $\hat{Q}(s,a)$), and subsequently updating the global solution ($V^{\bf g}(s)$ and/or $Q^{\bf g}(s,a)$ and/or $\pi^{\bf g}(a|s)$) based on the local solution. In Alg. \ref{alg_frap}, the updates of the local solution happen in lines 19 and 29, while the update of the global solution (when used) occurs in line 7. The main message of this section is that we can write both types of updates, whether it concerns updates of nodes in a planning tree or updates of a global policy network, as {\it gradient descent} updates on a particular {\it loss function}. We hope this provides further insight in the similarity between planning and learning, since planning updates on a tree/graph can usually be written as tabular learning updates with a particular learning rate. 

We will first introduce our general notation. A loss function is denoted by $\mathcal{L}(\theta)$, where $\theta$ denotes the parameters to be updated. In case of a tabular solution, the parameters are simply the individual entries in the table (like $Q^{\bf l}(s,a))$) (see Sec. \ref{sec_solution} and Table \ref{table_solution_types} for a summary of notation), and we will therefore not explicitly add a subscript $\theta$. When we have specified a solution and a loss function, the parameters can be updated based on gradient descent, with update rule: 

\begin{equation}
\theta  \gets \theta - \eta \cdot \nabla_\theta \mathcal{L}(\theta), \label{eq_gradient_descent}
\end{equation}

\noindent where $\eta \in \mathbb{R}^+$ is a learning rate. We will first show which loss function and update rules are common in updating of the local solution, and subsequently discuss how they reappear in updates of the global solution based on the local solution. An overview of common loss functions and update rules is provided in Table \ref{table_update_types}, which we will now discuss in more detail.

\paragraph{Local solution update} We will here focus on the update of state-action values $Q^{\bf l}(s,a)$ (Alg. \ref{alg_frap}, line 29), but the same principles apply to state value updating (Alg. \ref{alg_frap}, line 19). We therefore want to specify an update of $Q^{\bf l}(s,a)$ based on a new back-up value $\hat{Q}(s,a)$. A classic choice of loss function for continuous values is the {\it squared loss}, given by: 

\begin{equation}
\mathcal{L}\big(Q^{\bf l}(s,a)|s,a \big) = \frac{1}{2} \big[ \hat{Q}(s,a) - Q^{\bf l}(s,a)  \big]^2. \label{eq_squared_loss}
\end{equation}

\noindent Differentiating this loss with respect to $Q^{\bf l}(s,a)$ and plugging it into Eq. \ref{eq_gradient_descent} (where $Q^{\bf l}(s,a)$ are the parameters) gives the well-known {\it tabular learning rule}: 

\begin{equation}
Q^{\bf l}(s,a) \gets Q^{\bf l}(s,a) + \eta \cdot \big[ \hat{Q}(s,a) -  Q^{\bf l}(s,a) \big].  \label{eq_tabular_learning_rule}
\end{equation}

\noindent Intuitively, we move our estimate $Q^{\bf l}(s,a)$ a bit in the direction of our new back-up value $\hat{Q}(s,a)$. In the tabular case, $\eta$ is therefore restricted to $[0,1]$. Most planning algorithms use special cases of the above update rule. A first common choice is to set $\eta = 1.0$, which gives the {\it replace update}:

\begin{equation}
Q^{\bf l}(s,a) \gets \hat{Q}(s,a).
\end{equation}

\noindent This update completely overwrites the estimate in the local solution by the new back-up value. This is the typical approach in heuristic planning \citep{hart1968formal,nilsson1971problem,hansen2001lao}, where an admissible heuristic often ensures that our new estimate (from a deeper unfolding of the planning tree) provides a better informed estimate than the previous estimate. Although one would typically not think of such a replace update as a gradient-based approach, these updates are in fact all connected.

When we do not have a good heuristic available (and we therefore need to bootstrap from a learned value function or use deep roll-outs to estimate the cumulative reward), estimates of different depths may have different reliability (known as the {\it bias-variance trade-off}) \citep{sutton2018reinforcement}. We may for example equally weight the contribution of estimates of different depths, which we will call an {\it averaging update} (which uses $\eta = \frac{1}{n}$, where $n$ denotes the number of trials/back-up estimates for the node):

\begin{equation}
Q^{\bf l}(s,a) \gets Q^{\bf l}(s,a) + \frac{1}{n}  \cdot [ \hat{Q}(s,a) -  Q^{\bf l}(s,a) ]  
\end{equation} 

This is for example used in MCTS implementations that use bootstrapping instead of rollouts \citep{silver2018general}.

While the above update gives the value estimate from each trial equal weight, we may also make the contribution of a trial estimate dependent on the depth of the trial, as is for example done in {\it elegibility traces} \citep{sutton2018reinforcement,schulman2016high}. In this case, we essentially set $\eta = (1-\lambda) \cdot \lambda^{(d-1)}$, where $\lambda \in [0,1]$ is the exponential decay and $d$ is the length of the trace on which we update. More sophisticated reweighting schemes of the targets of different trials are possible as well \citep{munos2016safe}, for example based on the {\it uncertainty} of the estimate at each depth \citep{buckman2018sample}. In short, the local solution may combine value estimates from different trials (with different depths) in numerous ways, as summarized in the top part of Table \ref{table_update_types}.

\begin{figure}[!t]
  \centering
      \includegraphics[width = 0.9\textwidth]{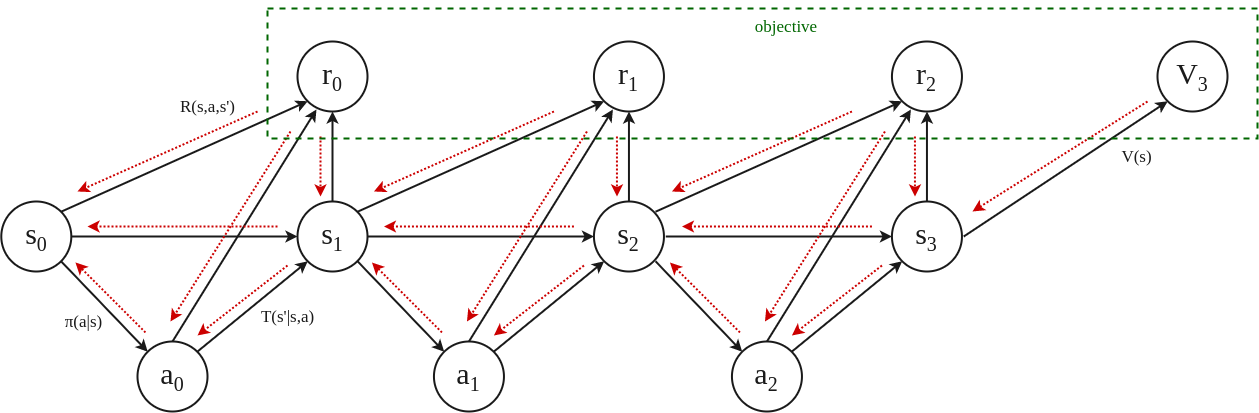}
  \caption{Illustration of gradient-based planning. When we have access to a differentiable transition function $\mathcal{T}(s'|s,a)$ and differentiable reward function $\mathcal{R}(s,a,s')$, and we also specify a differentiable policy $\pi_\theta(a|s)$, then a single trial generates a fully differentiable computational graph. The figure shows an example graph for a trial of depth 3. The black arrows show the forward passes through the policy, dynamics function and rewards function. In the example, we also bootstrap from a differentiable (learned) value function, but this can also be omitted. We may then update the policy parameters by directly differentiating the cumulative reward (objective, green box) with respect to the policy parameters, effectively summing the gradients over all backwards path indicated by the red dotted lines.}
    \label{fig_value_gradients}
\end{figure}

\begin{sidewaystable}
\footnotesize
\centering
\caption{Overview of common loss functions and update rules. {\bf Top}: Local update, where we use back-up values $\hat{V}(s)$ and/or $\hat{Q}(s,a)$ to update the local solution $V^{\bf l}(s)$ and/or $Q^{\bf l}(s,a)$. The special cases of replace update and average update are explicitly shown. {\bf Bottom}: Global update, where we use the local solution estimates $V^{\bf l}(s)$ and/or $Q^{\bf l}(s,a)$ to update global (approximate) solutions $V^{\bf g}_\theta(s)$, $Q^{\bf g}_\theta(s,a)$ and/or $\pi^{\bf g}_\theta(a|s)$. Parameters of the global solution are denoted by $\theta$ (when the global value solution is tabular each $\theta$ in the table can be read as $Q^{\bf g}(s,a)$) . Note that the table illustrates some characteristic examples, but other losses and update rules are possible. $\hat{Q}_d(s,a)$ denotes an estimate from a trial of depth $d$.} \label{table_update_types}
\begin{center}
\begin{tabular}{p{4.5cm} p{7cm} p{7.5cm} }
 & {\bf Loss} & {\bf Update} \\
\toprule
{\bf Local update} & \\
\quad {\it Value} & \\
\quad \quad Squared loss & $\mathcal{L}(Q^{\bf l}(s,a)|s,a) = \frac{1}{2} \big(\hat{Q}(s,a) - Q^{\bf l}(s,a) \big)^2$ & $Q^{\bf l}(s,a) \gets Q^{\bf l}(s,a) + \eta \cdot [ \hat{Q}(s,a) -  Q^{\bf l}(s,a) ]$  \\

\quad \quad \quad Replace update ($\eta = 1$) & & $Q^{\bf l}(s,a) \gets  \hat{Q}(s,a) $  \\
\quad \quad \quad Average update ($\eta = \frac{1}{n}$) & & $Q^{\bf l}(s,a) \gets Q^{\bf l}(s,a) + \frac{1}{n}  \cdot [ \hat{Q}(s,a) -  Q^{\bf l}(s,a) ]  $  \\
\quad \quad \quad Eligibility update & & $Q^{\bf l}(s,a) \gets Q^{\bf l}(s,a) + (1-\lambda) \cdot \lambda^{(d-1)} \cdot [ \hat{Q}_d(s,a) -  Q^{\bf l}(s,a) ]  $  \\

\midrule
{\bf Global update} & \\
\quad {\it Value} & \\
\quad \quad Squared loss & $\mathcal{L}(\theta|s,a) = \frac{1}{2} \big( Q^{\bf l}(s,a) - Q^{\bf g}_\theta(s,a)  \big)^2$ & $\theta \gets \theta + \eta \cdot [ Q^{\bf l}(s,a) - Q^{\bf g}_\theta(s,a) ] \cdot  \nabla_\theta Q^{\bf g}_\theta(s,a) $ \\
\quad \quad Cross-entropy softmax loss & $\mathcal{L}(\theta|s) = - \texttt{softmax}(Q^{\bf l}(s,{\bf a}))^T \cdot \log \texttt{softmax}(Q^{\bf g}_\theta(s,{\bf a}))$  & $\theta \gets \theta + \eta \cdot \nabla_\theta [\texttt{softmax}(Q^{\bf l}(s, {\bf a}))^T \cdot \log \texttt{softmax}((Q^{\bf g}_\theta(s,{\bf a}))] $ \\

\quad {\it Policy} & \\
\quad \quad Policy gradient & $\mathcal{L}(\theta|s,a) = - \ln \pi^{\bf g}_\theta(a|s) \cdot Q^{\bf l}(s,a)   $& $\theta \gets \theta + \eta \cdot \frac{Q^{\bf l}(s,a)} {\pi^{\bf g}_\theta(a|s)}  \cdot  \nabla_\theta \pi^{\bf g}_\theta(a|s) $ \\

\quad \quad Determ. policy gradient & $\mathcal{L}(\theta|s,a) = - Q^{\bf g}_\psi(s,\pi^{\bf g}_\theta(a|s)) $ \quad ($Q^{\bf g}_\psi$ trained on $Q^{\bf l}$ ) & $\theta \gets \theta + \eta \cdot \nabla_a Q^{\bf g}_\psi(s,a) \cdot \nabla_\theta \pi^{\bf g}_\theta(a|s) $ \\
\quad \quad Value gradient & $\mathcal{L}(\theta|s) = - V^{\bf l}(s) $ & $\theta \gets \theta + \eta \cdot  \nabla_\theta V^{\bf l}(s) $ \qquad \qquad \qquad (Fig. \ref{fig_value_gradients}) \\
\quad \quad Cross-entropy loss & $\mathcal{L}(\theta|s) = \sum_{a \in \mathcal{A}} \ln \pi_\theta(a|s_t) \big( \frac{n^{\bf l}(s_t,a)}{\sum_{b \in \mathcal{A}} n^{\bf l}(s_t,b)} \big) $& $\theta \gets \theta - \eta \cdot \sum_{a \in \mathcal{A}} \big( \frac{n^{\bf l}(s_t,a)}{\sum_{b \in \mathcal{A}} n^{\bf l}(s_t,b)} \big) \cdot \frac{1} {\pi^{\bf g}_\theta(a|s)}  \cdot  \nabla_\theta \pi^{\bf g}_\theta(a|s)$ \\
\bottomrule
\end{tabular}
\end{center}
\end{sidewaystable}

\paragraph{Global solution update} When our algorithm uses a global solution, we next need to update this global solution ($V^{\bf g}$ and/or $Q^{\bf g}$ and/or $\pi^{\bf g}$) based on the estimates from our local solution ($V^{\bf l}$ and/or $Q^{\bf l}$) (Alg. \ref{alg_frap}, line 7). For a value-based solution that is {\it tabular}, we typically use the same squared loss (Eq. \ref{eq_squared_loss}), which leads to the global tabular update rule $Q^{\bf g}(s,a) \gets Q^{\bf g}(s,a) + \eta \cdot [ Q^{\bf l}(s,a) - Q^{\bf g}(s,a) ]$, which exactly resembles the local version (Eq. \ref{eq_tabular_learning_rule}), apart from the fact that we now update $ Q^{\bf g}(s,a)$, while $ Q^{\bf l}(s,a)$ has the role of target. This approach is the basis under all tabular RL methods \citep{sutton2018reinforcement}. (For (model-free) RL approaches that directly update the global solution after a single trial, we may also imagine the local solution does not exist, and we directly update the global solution from the back-up estimates). 

We will therefore primarily focus on the function approximation setting, where we update a global approximate representation parametrized by $\theta$. Table \ref{table_update_types} shows some example loss functions and update rules that appear in this case. The most important point to note is that there are many ways in which we may combine a local estimate, such as $Q^{\bf l}(s,a)$, and the global solution, such as $Q^{\bf g}(s,a)$ or $\pi^{\bf g}(a|s)$, in a loss function. For value-based updating, we may use the squared loss, but other options are possible as well, like a cross-entropy loss over the softmax of the Q-values returned from planning (the local solution) and the softmax of the Q-values from a global neural network approximation \citep{hamrick2020combining}. For policy-based updating, well-known examples include the {\it policy gradient} \citep{williams1992simple,sutton2000policy,sutton2018reinforcement} and {\it deterministic policy gradient} \citep{silver2014deterministic,lillicrap2015continuous} loss functions. Again, other options have been successful as well, such as a cross-entropy loss between the normalized visitations counts at the root of an MCTS (part of the local solution) and a global policy network, as for example used by AlphaZero  \citep{silver2017mastering}. In short, various objectives are possible (and more may be discovered), as long as minimization of the objective moves our global solution in the right direction (based on the obtained information from the trial). 

An important other class of approaches is {\it gradient-based planning}, also known as {\it value gradients} \citep{fairbank2012value,heess2015learning}. These approaches require a (known or learned) differentiable transition and reward model (and a differentiable value function when we also include bootstrapping). When we also specify a differentiable policy, then each trial generates a fully differentiable graph, in which we can directly differentiate the cumulative reward with respect to the policy parameters. This idea is illustrated in Fig. \ref{fig_value_gradients}, where we aggregate over all gradient paths in the graph (red dotted lines). Gradient-based planning is popular in the robotics and control community  \citep{anderson2007optimal,todorov2005generalized,deisenroth2011pilco}, where dynamics functions are relatively smooth and differentiable, although the idea can also be applied with discrete states \citep{wu2017scalable}. 

Table \ref{table_update_types} summarizes some of the common loss functions we discussed. The examples in the table all have analytical gradients, but otherwise we may always use finite differencing to numerically estimate the gradient of an objective. The learning rate in these update equations is typically tuned to a specific value (or decay scheme), although there are more sophisticated approaches that bound the step size, such as proximal policy optimization (PPO) \citep{schulman2017proximal}. Moreover, we did not discuss gradient-free updating of a global solution, because these algorithms typically do not exploit MDP-specific knowledge (i.e, they do not construct and back-up value estimates at states throughout the MDP, but only sample the objective function based on traces from the root). However, we do note that gradient-free black-box optimization can also be successful in MDP optimization, as for example show for evolutionary strategies \citet{moriarty1999evolutionary,whiteson2006evolutionary,salimans2017evolution}, simulated annealing \citep{atiya2003reinforcement} and the cross-entropy method \citet{mannor2003cross}.

This concludes our discussion of the dimensions in the framework. An overview of all considerations and their possible choices is shown in Table \ref{table_framework}, while Algorithm \ref{alg_frap} shows how all these considerations piece together in a general algorithmic framework. To illustrate the validity of the framework, the next section will analyze a variety of planning and RL methods along the framework dimensions. 

\begin{sidewaystable}[p]
\centering
\scriptsize
\caption{Comparison of algorithms (columns) along the framework dimensions (rows). Blue, red and green colour denote planning, model-free RL and model-based RL algorithms, respectively (although Value Iteration is technically model-based RL under our definitions in Sec. \ref{sec_problem_def}, we still list it as first entry since it is a core algorithm). All methods that use a global solution also use a local solution (which we did not explicitly write in the table). Regarding action selection, when applicable we discriminate {\it before frontier} (BF) action selection, {\it after frontier} (AF) action selection, and {\it next root} (NR) action selection. When the squared loss is written between brackets, it means that the algorithm uses a direct tabular update rule and the squared loss is therefore never explicitly part of the algorithm. NN = neural network, GP = Gaussian Process, PPO = Proximal Policy Optimization \citep{schulman2017proximal}.} \label{table_overview}
\begin{tabular}{ p{1.5cm} p{2.5cm} | P{2.0cm} P{2.0cm} P{2.0cm} P{2.0cm} P{2.0cm} P{2.0cm} P{2.0cm}}
\toprule
\bfseries{Dimension} \newline  &  \bfseries{Consideration} &  Value iteration \citep{bellman1966dynamic} & \cellcolor{blue!25} LAO$^\star$ \citep{hansen2001lao} & \cellcolor{blue!25} Labeled RTDP \citep{bonet2003labeled} & \cellcolor{blue!25} Monte Carlo search \citep{tesauro1997online} & \cellcolor{blue!25} MCTS \citep{kocsis2006bandit} & \cellcolor{red!25}  Q-learning \citep{watkins1992q} & \cellcolor{red!25} TD($\lambda$) \citep{sutton2018reinforcement} \newline \\
  \hline			
 & & \\

MDP access &  & Settable descriptive & Settable descriptive & Settable descriptive & Settable generative & Settable generative & Resettable generative & Resettable generative \\

& & \\

Solution & - Coverage & Global & Local & Local & Local & Local & Global & Global  \\ 
 & - Type & $V(s)$ & $V(s)$ & $V(s)$ & $Q(s,a)$ & $Q(s,a)$ & $Q(s,a)$ & $V(s)$ \\  
 & - Method & Tabular & Tabular & Tabular & Tabular & Tabular & Tabular & Tabular \\ 
 & - Initialization & Uniform & Heuristic & Heuristic & Uniform & Optimistic & Uniform & Uniform \\ 

& & \\ 
 
Root & - Selection  & Ordered & Forward sampling & Forward sampling & Forward sampling & Forward sampling & Forward sampling & Forward sampling \\

& & \\

Budget & - \# trials per root & up to $| \mathcal{A} | \cdot | \mathcal{S} |$  & till convergence & up to $| \mathcal{A} | \cdot | \mathcal{S} |$  & $n$ & $n$ &  1 & $d_{\max}$ \\

 & - Depth & 1 & $1..n$ & 1 & $\infty$ & $\infty$ & 1  & $1..d_{\max}$ \\

& & \\

Selection & - Next action & Ordered & BF: Greedy, AF: Ordered, NR: Greedy & BF: Greedy, AF: Ordered, NR: Greedy  & BF: Ordered AF: Baseline & BF: Uncertainty AF: Baseline NR: Greedy & Random pert. & Random pert. \\
 & - Next state & Ordered & Ordered & Sample & Sample & Sample & Sample & Sample \\
& & \\

Bootstrap& - Location  & State & State & State & - & - & State-action & State \\
& - Type  & Learned & Heuristic & Heuristic & - & - & Learned & Learned \\

& & \\

Back-up & - Back-up policy & Greedy/max & Greedy/max & Greedy/max & On-policy & On-policy & Greedy/max & On-policy \\
 & - Policy exp. & - & - & - & Sample & Sample & - & Sample \\
 & - Dynamics exp. & Expected & Expected & Expected & Sample & Sample & Sample & Sample \\
 & - Add. back-ups & - & Convergence label & Convergence label & - & Counts & - & - \\

& & \\

Update & - Loss  & (Squared) & (Squared) & (Squared) & (Squared) & (Squared) & (Squared) & (Squared) \\ 
& - Update type & Replace ($\eta=1.0$) & Replace ($\eta=1.0$) & Replace ($\eta=1.0$) & Average ($\eta=1/n$) & Average ($\eta=1/n$) & Fixed step & Eligibility  \\ 
\bottomrule

\end{tabular}

\end{sidewaystable}

\begin{sidewaystable}[p]
\ContinuedFloat
\centering
\scriptsize
\caption{Continued.}
\begin{tabular}{ p{1.5cm} p{2.5cm} | P{2.0cm} P{2.0cm} P{2.0cm} P{2.0cm} P{2.0cm} P{2.0cm} P{2.0cm}}
\toprule
\bfseries{Dimension} \newline  &  \bfseries{Consideration} & \cellcolor{red!25} REINFORCE \citep{williams1992simple} & \cellcolor{red!25} DQN \citep{mnih2015human} & \cellcolor{green!25} Prioritized sweeping \citep{moore1993prioritized} & \cellcolor{green!25} Dyna \citep{sutton1990integrated} & \cellcolor{green!25} PILCO \citep{deisenroth2011pilco} & \cellcolor{green!25} AlphaGo \citep{silver2017mastering} & \cellcolor{green!25} Go-Explore (policy-based) \citep{ecoffet2021first} \newline \\
  \hline			
 & & \\

MDP access &  & Resettable generative  & Resettable generative & Resettable generative & Resettable generative & Resettable generative & Settable generative & Resettable generative \\

& & & & & & & &  \\

Solution & - Coverage & Global & Global & Global & Global & Global & Global & Global  \\ 
 & - Type & $\pi(a|s)$ & $Q(s,a)$ & $Q(s,a)$ & $Q(s,a)$ & $\pi(a|s)$ & $\pi(a|s)$, $V(s)$ & $\pi(a|s,g)$, $V(s)$ \\  
 & - Method & Tabular & Approximate (NN) & Tabular & Tabular & Approximate (GP) & Approximate (NN) & Approximate (NN)  \\ 
 & - Initialization & Uniform & Random & Uniform & Uniform & Random & Random & Random \\ 

& & \\ 
 
Root & - Selection  & Forward & Forward & Forward + backward & Forward + visited states & Forward & Forward & Forward  \\

& & \\

Budget & - \# trials per root & 1 & 1 & 1 & 1 & 1 & 1600 & $d_{\max}$  \\

 & - Depth & $\infty$ & 1 & 1 & 1 & $\infty$ & MCTS: $1..n$ \newline NR: $\infty$ & $1..d_{\max}$ \\

& & \\

Selection & - Next action & Rand. pert. (stoch. policy) & Rand. pert. ($\epsilon$-greedy) & State-based (novelty) & State-based (novelty) + Mean pert. (Boltzmann) & Rand. pert. (stoch. policy) & BF/AF: Uncertainty NR: Rand. pert. & BF: Novelty + Mean pert. (entropy), AF: Rand. pert.\\
 & - Next state & Sample & Sample & Sample & Sample & Sample & Sample & Sample \\
& & \\

Bootstrap& - Location  & -  & State-action  & State-action & State-action & - & State & State \\
& - Type  & - & Learned & Learned & Learned & - & Learned & Learned \\

& & \\

Back-up & - Back-up policy & On-policy & Max/greedy & Max/greedy &  On-policy & On-policy & On-policy & On-policy \\
 & - Policy exp. & Sample & - & Max  &  Sample & Sample & Sample & Sample \\
 & - Dynamics exp. & Sample & Sample & Expected & Sample & Sample & Sample & Sample \\
 & - Add. back-ups & - & - & Priorities, counts & Counts & Uncertainty & Counts & Counts \\

& & \\

Update & - Loss  & Policy gradient  & Squared & (Squared) & (Squared) & Value gradient & Cross-entropy (policy) + squared (value) & Policy gradient (PPO) + squared (value) \\ 
& - Learning rate  & Fixed step & Fixed step & Fixed step & Fixed step & Fixed step & Local: Average Global: fixed step & Local: eligibility Global: adaptive \\ 
\bottomrule

\end{tabular}

\end{sidewaystable}

\section{Comparison of algorithms} \label{sec_comparison}
Having discussed all the dimensions of the framework, we will now zoom out and reflect on its use and potential implications. The main point of our framework is that MDP planning and reinforcement learning algorithms occupy the same solution space. To illustrate this idea, Table \ref{table_overview} shows for a range of well-known planning (blue), model-free RL (red) and model-based RL (green) algorithms the choices they make on the dimensions of the framework. The list is of course not complete (we could have included any other preferred algorithm), but the table illustrates that the framework is applicable to a wide range of algorithms. 

A first observation from the table is that it reads like a patchwork. On most dimensions the same decisions appear in both the planning and reinforcement learning literature, showing that both fields actually have quite some overlap in developed methodology. For example, the depth and back-up schemes of MCTS \citep{kocsis2006bandit} and REINFORCE \citep{williams1992simple} are exactly the same, but they differ in their solution coverage (MCTS only uses a local solution, REINFORCE updates a global solution after every trial) and exploration method. Such comparisons provide insight into the overlap and differences between various approaches. 

The second observation of the table is therefore that {\it all algorithms have to make a decision on each dimension}. Therefore, even though we often do not consciously consider each of the dimensions when we come up with a new algorithm, we are still implicitly making a decision on each of them. The framework could thereby potentially help to structure the design of new algorithms, by consciously walking along the dimensions of the framework. It also shows what we should actually report about an algorithm to fully characterize it.

There is one deeper connection between planning and tabular reinforcement learning we have not discussed yet. In our framework, we treated the back-up estimates generated from a single model-free RL trial as a local solution. This increases consistency (i.e., allows for the pseudocode of Alg. \ref{alg_frap}), but we could also view model-free RL as a direct update of the global solution based on the back-up estimate (i.e., skip the local solution). With this view we see another relation between common planning and tabular learning algorithms, such as MCTS (planning) and Monte Carlo reinforcement learning (MCRL). Both these algorithms sample trials and compute back-up estimates in the same way, but MCTS writes these to a local tabular solution (with learning rate $\eta = \frac{1}{n}$), while MCRL writes these to a global tabular solution (with fixed learning rate $\eta$). These algorithms from different research fields are therefore strongly connected, not only in their back-up, but also in their update schemes. 

We will briefly emphasize elements of the framework, or possible combinations of choices, that could deserve extra attention. First of all, the main success of reinforcement learning originates from its use of global, approximate representations \citep{silver2017mastering,ecoffet2021first}, for example in the form of deep neural networks. These approximate representations allow for generalization between similar states, and planning researchers may therefore want to emphasize global solution representations in their algorithms. The other way around, a main part of the success of planning literature comes from the stability and guarantees of building local, tabular solutions. Combinations of both approaches show state-of-the-art results \citep{silver2017mastering,levine2014learning,hamrick2020combining}, and each illustrate that we can be very creative in the way learned global solutions can guide new planning iterations, and the way planning output may influence the global solution and/or action selection. Important research questions are therefore how action selection within a trial can be influenced by the global solution (Alg. \ref{alg_frap}, line 16), how a local solution should influence the global solution (i.e., variants of loss functions, Alg. \ref{alg_frap}, line 7), and how we may adaptively assign planning budgets per root state (Alg. \ref{alg_frap}, line 5).

Another important direction for cross-pollination is the study of {\it global, approximate frontiers}. On the one hand, planning research has extensively studied the benefit of local, tabular frontiers, a crucial idea which has bee ignored in most RL literature. On the other hand, tabular frontiers do not scale to high-dimensional problems, and in these cases we need to track some kind of global approximate frontier, as studied in intrinsically motivated goal exploration processes \citep{colas2020intrinsically}. Initial results in this direction are for example provided by \citet{pere2018unsupervised, ecoffet2021first}, but there appears to be much remaining research in this field. Getting back to the previous point, we also believe semi-parametric memory and episodic memory \citep{blundell2016model,pritzel2017neural} may play a big role for global approximate solutions, for example to ensure we can directly get back to a recently discovered interesting state. 

A third interesting direction is a stronger emphasis on the idea of backward search (planning terminology) or prioritized sweeping (RL terminology). In both communities, backward search has received considerable less attention than forward search, while backward approaches are crucial to spread acquired information efficiently over a (global) state space (by setting root states in a smarter way, see Sec. \ref{sec_set_root node}). The major bottleneck seems the necessity of a {\it reverse} model (which state-actions may lead to a particular state), which is often available in smaller, tabular problems, but not in large complex problems where we only have a simulator or real world interaction available. However, we may learn an approximate reverse model from data, which could bring these powerful ideas back into the picture. Initial (promising) results in this direction are provided by \citet{edwards2018forward,agostinelli2019solving,corneil2018efficient}. 

In summary, the framework for reinforcement learning and planning (FRAP), as presented in this paper, shows that both planning and reinforcement learning algorithms share the same algorithmic space. This provides a common language for researchers from both fields, and may help inspire future research (for example by cross-pollination). Finally, we hope the paper also serves an educational purpose, for researchers from one field that enter into the other, but particularly for students, as a systematic way to think about the decisions that need to be made in a planning or reinforcement learning algorithm, and as a way to integrate algorithms that are often presented in disjoint courses.

\bibliographystyle{apalike}
\bibliography{overview}

\end{document}